\documentclass[runningheads]{llncs}
\usepackage{appendix}

\usepackage[utf8]{inputenc}
\usepackage{array}
\usepackage{wrapfig}
\usepackage{multirow}
\usepackage{tabu}

\usepackage{amsfonts}
\usepackage{booktabs}
\usepackage{siunitx}

\usepackage{threeparttable,booktabs}

\usepackage{geometry}
\usepackage{changepage}
\usepackage{enumitem}
\usepackage[sort,numbers,sectionbib]{natbib}

\makeatletter
\newif\if@restonecol
\makeatother

\usepackage[ruled,vlined]{algorithm2e}

\usepackage{amssymb}
\usepackage{pifont}
\newcommand{\cmark}{\color{blue}\ding{51}}%
\newcommand{\xmark}{\color{red}\ding{55}}%
\newcommand{\bdot}{\color{black}\ding{108}}%
\newcommand{\rdot}{\color{red}\ding{108}}%

\usepackage{bm}
\usepackage{tikz,pgfplots,pgfplotstable}
\usetikzlibrary{pgfplots.groupplots}
\usetikzlibrary{fit,arrows, backgrounds, calc, hobby, positioning}
\usetikzlibrary{shapes.geometric}
\begin{document}

\titlerunning{INCAD for Streaming Data}
\authorrunning{S. Guggilam et al. }

\title{Integrated Clustering and Anomaly Detection (INCAD) for Streaming Data}
\author{
Sreelekha Guggilam\inst{1}
\and
Syed M. A. Zaidi\inst{2}
\and
    Varun Chandola\inst{1,2}
\and
  Abani K. Patra\inst{1}
}
\institute{Computational Data Science \& Eng., \and
Computer Science \& Eng.,
University at Buffalo, State University of New York (SUNY)
\email{\{sreelekh,szaidi2,chandola,abani\}@buffalo.edu}
}


\maketitle
\begin{abstract}
Most current clustering based anomaly detection methods use scoring schema and thresholds to classify anomalies. These methods are often tailored to target specific data sets with ``known'' number of clusters. The paper provides a streaming clustering and anomaly detection algorithm that does not require strict arbitrary thresholds on the anomaly scores or knowledge of the number of clusters while performing probabilistic anomaly detection and clustering simultaneously. This ensures that the cluster formation is not impacted by the presence of anomalous data, thereby leading to more reliable definition of {\it``normal vs abnormal''} behavior. The motivations behind developing the INCAD model \cite{INCAD} and the path that leads to the streaming model is discussed.
\end{abstract}


\section{Introduction}\label{sec1}

Anomaly detection heavily depends on the definitions of expected and anomalous behaviors \cite{kruegel2003anomaly,garcia2009anomaly,Jiang:2015}. In most real systems, observed system behavior typically forms natural clusters whereas anomalous behavior either forms a small cluster or is weakly associated with the natural clusters. Under such assumptions, clustering based anomaly detection methods form a natural choice \cite{chan2003machine,eskin2002geometric,he2003discovering} but have several limitations. 

Firstly, clustering based methods usually require baseline assumptions that are often conjectures and generalizing them is not always trivial. This leads to inaccurate choices for model parameters such as the number of clusters or the thresholds that are required to classify anomalies. Score based models have thresholds that are often based on data/user preference. Such assumptions result in models that are susceptible to modeler's bias and possible over-fitting.

Secondly, setting the number of clusters has additional challenges when dealing with streaming data, where new behavior could emerge and form new clusters.   Non-stationarity is inherent as data evolves over time. Moreover, the data distribution of a stream changes over time due to changes in environment, trends or other unforeseen factors~\cite{gama2014survey, jiang2006research}. This leads to a phenomenon called \textit{concept drift}, due to which an anomaly detection algorithm cannot assume any fixed  distribution for data streams. Thus, there arises a need for a definition of an anomaly that is dynamically adapted.

Thirdly, when anomaly detection is performed post clustering \cite{amer2012nearest,fu2005similarity}, the presence of anomalies gives a  skewed (usually slight) definition of traditional/normal behavior.  However, since the existence of anomalies impacts the clustering as well as the definition of the `{\it normal}'\footnote{Non-anomalous behavior is described as \lq\lq normal" behavior. Should not be confused with Gaussian/Normal distribution.} behavior, it seems counter-intuitive to classify anomalies based on such definitions\footnote{Clustering and defining \lq\lq normal/traditional" behavior in presence of anomalies develop in skewed and inconsistent results.}.To avoid this, simultaneous clustering and anomaly detection needs to be performed.
\begin{table*}
\caption{Comparison with other anomaly detection methods} 
\label{table1}
\resizebox{\textwidth}{!}{
\begin{tabular}{lccccccc}
\hline\hline 
    & Neural Networks & LOF & KNN & Kmeans -- & Kernel Function Based & Gaussian Model Based & INCAD    \\
  \hline  
Clustering Based                & \xmark  & \xmark  & \cmark       & \cmark                   & \xmark                   & \xmark     & \cmark \\
Multi-dimension                 & \cmark & \xmark  & \cmark       & \cmark                   & \cmark                  & \cmark    & \cmark \\
Unsupervised                    & \xmark  & \cmark & \cmark       & \cmark                   & \cmark                  & \cmark    & \cmark \\
Non-parametric                  & \xmark  & \xmark  & \xmark        & \xmark                    & \cmark                  & \xmark     & \cmark \\
Adaptable to streaming settings & \xmark  & \xmark  & \xmark        & \xmark                    & \xmark                   & \xmark     & \cmark \\
Adaptive thresholds             & \xmark  & \xmark  & \xmark        & \xmark                    & \xmark                   & \xmark     & \cmark \\
Probabilistic scoring           & \xmark  & \xmark  & \xmark        & \xmark                    & \cmark                  & \xmark     & \cmark\\
\hline
\end{tabular}}
\end{table*}

In addition to the above challenges, extending these assumptions to the streaming context leads to a whole new set of challenges. Many supervised~\cite{chandola2009anomaly, gornitz2013toward} and unsupervised anomaly detection techniques~\cite{bay2003mining, chandola2009anomaly, goldstein2016comparative, he2003discovering} are offline learning methods that require the full data set in advance for data mining which makes them unsuitable for real-time streaming data. Although supervised anomaly detection techniques may be effective in yielding good results, they are typically unsuitable for anomaly detection in streaming data~\cite{gornitz2013toward}.
We propose a method called  Integrated Clustering and Anomaly Detection (INCAD), that couples Bayesian non-parametric modeling and extreme value theory to simultaneously perform clustering and anomaly detection. Table \ref{table1} summarizes the properties of INCAD vs other strategies for anomaly detection. The primary contributions of the paper are as follows:
\begin{enumerate}
    \item \textbf{Generalized anomaly definition with adaptive interpretation} The model definition of an anomaly has dynamic interpretation allowing anomalous behaviors to evolve into normal behaviors and vice versa. This definition not only evolves the number of clusters with an incoming stream of data (using non-parametric mixture models) but also helps evolve the classification of anomalies.
    \item \textbf{Combination of Bayesian non-parametric models and extreme value theory(EVT)} The novelty of the INCAD approach lies in blending extreme value theory and Bayesian non-parametric models. Non-parametric mixture models~\cite{Hjort:2010}, such as {\em Dirichlet Process Mixture Models} (DPMM)~\cite{Antoniak:1974, Rasmussen:2000, Teh:2006}, allow the number of components to vary and evolve during inference. While there has been limited work that has explored DPMM for the task of anomaly detection~\cite{Shotwell:2011,Varadarajan:2017}, they have not been shown to operate in a streaming mode or ignore online updates to the DPMM model. On the other hand, EVT gives the probability of a point being anomalous which has a more universal interpretation, in contrast to the scoring schema with user-defined thresholds. Although EVT's definition of anomalies is more adaptable for streaming data sets~\cite{Siffer:2017,al2016semi, french2019quantifying}, fitting an extreme value distribution (EVD) on a mixture of distributions or even multivariate distributions is challenging. This novel combination brings out the much-needed aspects in both the models.
    
    \item \textbf{Extension to streaming settings} The model is non-exchangeable which is an well suited   to capture the effect of the order of data input and utilize this dependency to develop streaming adaptation. 

    \item \textbf{Ability to handle complex data generative models} The model can be generalized to multivariate distributions and complex mixture models.
\end{enumerate}

\section{Motivation}\label{sec3}
\subsection{Assumptions on \textit{Anomalous} Behavior }
One of the key drivers in developing any model are the model assumptions. For INCAD model, we assume that the data has multiple {\em ``normal"} as well as {\em ``anomalous"} behaviors. These behaviors are dynamic with a tendency to evolve from {\em ``anomalous"} to {\em ``normal"} and vice versa. Each such behavior (normal/anomalous) forms a sub-population that can be represented using a cluster. These clusters are assumed to be generated from a family of  distributions whose cluster proportions and cluster parameters are generated from a non-parametric distribution. 

There are two distinct differences between normal and anomalous data that must be identified: (a) {\em ``Anomalous"} instances are different  from tail instances of {\em ``normal"} behavior and need to be distinguished from them. They are assumed to be generated from distributions that are different from {\em ``normal"} data. (b) The distributions for anomalous data result in relatively fewer instances in the observed set. 

As mentioned earlier, clustering based anomaly detection methods could be good candidates for monitoring such systems, but require the ability to allow the clustering to {\em evolve} with the streaming data, i.e., new clusters form, old clusters grow or split. Furthermore, we need the model to distinguish between anomalies and extremal values of the {\it ``normal''}.

Thus, non-parametric models that can accommodate infinite clusters are integrated with extreme value distributions that distinguish between anomalous and non-anomalous behaviors. 

We now describe the two ingredients that go into our model namely, mixture models and extensions of EVT.
\subsection{EVT and Generalized Pareto Distribution in Higher Dimensions}
Estimation of parameters for extreme value distributions in higher dimensions is complex. To overcome this challenge, \citet{Clifton:2014} proposed an extended version of the generalized Pareto distribution that is applicable for different multi-model and multivariate distributions. For a given data $X\in\mathbb{R}^n$  distributed as $f_X:X\rightarrow{}Y$ where $Y\in\mathbb{R}$ is the image of the pdf values of $X$, let $Y\in[0,y_{max}]$ be the range of $Y$ where $y_{max}=sup(f_X)$.
Let, 
\begin{equation}
    G_Y(y)=\int_{f_Y^{-1}([0,y])}^{} f_X(x)dx
\end{equation}
where $f_Y^{-1}:Y\rightarrow{}X$ is the pre-image of $f_X$ given by, $f_Y^{-1}([0,y])=\{x|f_X(x)\in[0,y]\}$. Then $G_Y$ is in the domain of attraction of generalized Pareto distribution (GPD) $G^e_Y$ for $y\in[0,u]$ as $u\rightarrow{}0$ given by,
\begin{eqnarray}
    G^e_Y(y)=\left\{
    \begin{array}{cc}
          1-(1-\xi(\frac{y-\nu}{\beta})^{-1/\xi}, &\xi\neq0\\
          1-exp(-\frac{y-\nu}{\beta}), &\xi=0
    \end{array}
    \right.
\end{eqnarray} 
where, $\nu, \beta$and $\xi$ are the location, scale and shape parameters of the GPD respectively.

\subsection{ Mixture Models}\label{sec31}
Mixture models assume that the data consists of sub-populations each generated from a different distribution. It can be used to study the properties of clusters using mixture distributions. In the classic version, when the number of clusters is known, finite mixture models are used with Dirichlet priors. However, when the number of latent clusters is unknown, one can extend finite mixture models to infinite mixture models like Dirichlet process mixture model (DPMM). In DPMM, the mixture distributions being sampled from a Dirichlet process (DP). 
DP can be viewed as a distribution over a family of distributions, that constitutes a base distribution $G_0$ which is a prior over the cluster parameters $\theta$ and positive scaling parameter $\alpha_{DP}$. $G$ is a Dirichlet process (denoted as $G\sim DP(G_0,\alpha_{DP})$) if $G$ is a random distribution with same support as  the base distribution $G_0$ and for any measurable finite partition of the support $A_1 \cup A_2 \cup \ldots \cup A_k$, we have
$(G(A_1), G(A_2), \ldots, G(A_k)) \sim $ Dir($\alpha_{DP} G_0(A_1),\ldots,\alpha_{DP} G_0(A_k)$). 

In order to learn the number of clusters from the data, Bayesian non-parametric (BNP) models are used. BNP models like DPMM assume an infinite number of clusters of which only a finite number  are populated. It brings forth a finesse in choosing the number of clusters while assuming a prior on the cluster assignments of the data. The prior is given by the Chinese restaurant process(CRP) which is defined analogously to the seating of $N$ customers who sequentially join tables in a Chinese restaurant. 
Here, the probability of the $n^{th}$ customer joining an existing table is proportional to the table size while the probability the customer forms a new table is always proportional to parameter $\alpha$, $\forall n\in{1,2,..,N}$. This results in a distribution over the set of all partitions of integers $1,2,..,N$. More formally, the distribution can be represented using the following probability function:
\begin{eqnarray}
P(z_n=k|z_{1:n-1})= \left\{ 
\begin{array}{cc}
 \frac{n_k}{n+\alpha-1} & n_k>0 \text{ (existing cluster)}  \\
  \frac{\alpha}{n+\alpha-1}   & n_k=0\text{ (new cluster)}
\end{array}
\right.
\end{eqnarray} 
where $z_i$ is the cluster assignment of the $i^{th}$ data point, $n_k$ is the size of the $k^{th}$ cluster , $\alpha>0$ is the concentration parameter. Large $\alpha$ values corresponds to an increased tendency of data points to form new clusters\footnote{Our modified model  targets this aspect of concentration parameter to generate the desired simultaneous clustering and anomaly detection.}.

\begin{figure}[htbp]
  \centering
  \begin{tikzpicture}{scale=0.8}
    \tikzstyle{main}=[circle, minimum size = 10mm, thick, draw =black!80, node distance = 16mm]
    \tikzstyle{mainobs}=[circle, minimum size = 10mm, thick, fill=black!10,draw =black!80, node distance = 16mm]
    \tikzstyle{connect}=[-latex, thick]
    \tikzstyle{box}=[rectangle, draw=black!100]
    \node[mainobs] (x) {$x_i$};
    \node[main] (z) [left=of x,xshift=-1em] {$z_i$};
    \node[main] (pi) [above=of z,xshift=-2em]{${\bm \pi}$};
    \node (alpha) [above=of pi]{$\alpha$};
    \node[main] (theta) [above=of x,xshift=-2.5em] {${\theta_k}$};
    \node (lambda) [above=of theta,xshift=2.5em] {$\lambda$};
    \node[main] (thetaa) [above=of x,xshift=2.5em] {${\theta^a_k}$};
    \node[main] (pia) [above=of z,xshift=2em] {${\bm \pi}^a$};
    \node (alphaa) [above=of pia] {$\alpha^a$};
    \node[main] (a) [below=of x,xshift=-4em,yshift=2em] {$a_i$};
    \node (gamma) [below=of a] {$\gamma$};
    \path
    (alpha) edge [connect] (pi)
    (lambda) edge [connect] (theta)
    (lambda) edge [connect] (thetaa)
    (alphaa) edge [connect] (pia)
    (pi) edge [connect] (z)
    (z) edge [connect] (x)
    (pia) edge [connect] (z)
    (gamma) edge [connect] (a)
    (a) edge [connect] (x)
    (a) edge [connect] (z)
    (theta) edge [connect] (x)
    (thetaa) edge [connect] (x);
    \node[rectangle, inner sep=4mm,draw=black!100, fit = (x) (z) (a)] (rect1) {};
    \node[rectangle, inner sep=2mm,draw=black!100, fit = (theta)] (rect2) {};
    \node[rectangle, inner sep=2mm,draw=black!100, fit = (thetaa)] (rect3) {};
    \node[anchor=south east] at (rect1.south east) {\footnotesize $N$};
    \node[anchor=south east] at (rect2.south east) {\footnotesize $\infty$};
    \node[anchor=south east] at (rect3.south east) {\footnotesize $\infty$};
  \end{tikzpicture}
  \caption{Graphical representation of the proposed INCAD model.}
  \label{fig:modelpgm}
\end{figure}
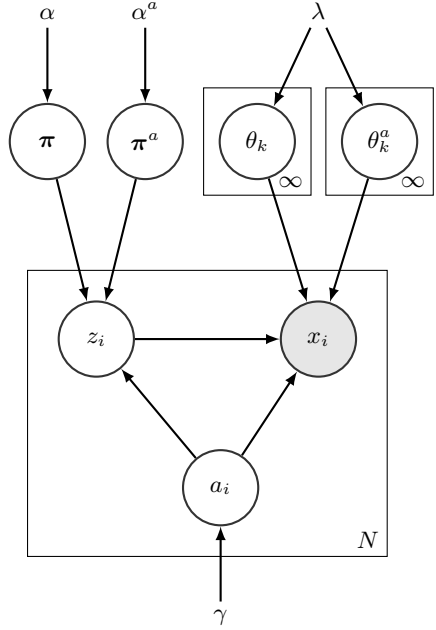

\section{Integrated Clustering and Anomaly Detection (INCAD)}
The proposed INCAD model's prior is essentially a modification of a Chinese Restaurant Process (CRP). The seating of a customer at the Chinese restaurant is dependent on an evaluation by a gatekeeper. The gatekeeper decides the ability of the customer to start a new table based on the customer's features relative to existing patrons. If the gatekeeper believes the customer stands out, the customer is assigned a higher concentration parameter $ \alpha $ that bumps up their chances of getting a new table.
The model was inspired by the work of \citet{blei2010distance} in their distance dependent CRP models. The integrated INCAD model defines a flexible concentration parameter $\alpha$ for each data point. The probabilities are given by:
\begin{eqnarray*}
    P(z_n|z_{1:n-1},\textbf{x})=\left\{
    \begin{array}{cc}
        \frac{n_k}{n+\alpha_2 -1} & , \ n_k>0 \ (existing \ cluster) \\
        \frac{\alpha_2}{n+\alpha_2 -1} & , \ n_k=0 \ (new \ cluster) \\ 
    \end{array}\right.
\end{eqnarray*}
where $z,n,n_k$ are described earlier and $\alpha_2=f(\alpha|x_n,\textbf{x},\textbf{z})$ is a function of a base concentration parameter $\alpha$. 
The function is chosen such that it is monotonic in $p(x_n)$ where, p is the probability of $x_i$ being in the tail of the mixture distribution.
In this paper, $f$ is given by, 
\begin{eqnarray*}
    f(\alpha|x_n,\textbf{x},\textbf{z})=\left\{\begin{array}{cc}
          \alpha &, if\ not\ in\ tail\\
           \alpha^* &,if\ in\ tail
    \end{array}\right.
\end{eqnarray*}
where, $\alpha^*=\frac{100}{1-p_n}$ and 
\begin{eqnarray}
    p_n=p(x_n)=\left\{\begin{array}{cc}
          \text{Probability of } x_n \text{ being anomalous,} & x_n \text{ in tail}\\
          0, &\text{otherwise}
    \end{array}\right.\label{eq:incadpi}
\end{eqnarray}
Traditional CRP not only mimics the partition structure of DPMM but also allows flexibility in partitioning differently for different data sets. However, CRP lacks the ability to distinguish anomalies from non-anomalous points. To set differential treatment for anomalous data, the CRP concentration parameter $\alpha$ is modified to be sensitive to anomalous instances and nudge them to form individual clusters. The tail points are clustered using this updated concentration parameter $\alpha_2$ which is designed to increase with increasing probability of a point being anomalous. This ensures that the probability of tail points forming individual clusters increases as they are further away from the rest of the data. 

\section{Choice of the Extreme Value Distribution}
The choice for $G^{EV}_0$, the base distribution for the anomalous cluster parameters, is the key for identifying anomalous instances. By choosing $G^{EV}_0$ as the extreme value counterpart of $G_0$, the model ensures that the anomalous clusters are statistically ``far'' from the normal clusters. However, as discussed earlier, not all distributions have a well-defined EVD counterpart. We first describe the inference strategy for the case where $G^{EV}_0$ exists. We then adapt this strategy for the scenario where $G^{EV}$ is not available. 

\subsection{Inference when $G^{EV}_0$ is available}
Inference for traditional DPMMs is computationally expensive, even when conjugate priors are used. MCMC based algorithms~\cite{Neal:2000} and variational techniques~\cite{Blei:2004} have been typically used for inference. Here we adopt the Gibbs sampling based MCMC method for conjugate priors (Algorithm \ref{alg:generator1}~\cite{Neal:2000}). This algorithm can be thought of as an extension of a Gibbs sampling-based method for a fixed mixture model, such that the cluster indicator $z_i$ can take values between $1$ and $K+1$, where $K$ is the current number of clusters. 

\subsubsection{Gibbs Sampling}
Though INCAD is based on DPMM, the model has an additional anomaly classification variable $a_.$ that determines the estimation of the rest of the parameters. In the Gibbs sampling algorithm for INCAD, the data points $\{x_i\}_{i=1}^N$ are observed and the number of clusters $K$, cluster indicators $\{z_i\}_{i=1}^N$ and anomaly classifiers $\{a_i\}_{i=1}^N$ are latent. Using Markov property and Bayes rule, we derive the posterior probabilities for $z_i\ \forall i\in {1,2,\dots, N}$ as:
\begin{eqnarray*}
P(z_i=k\vert x_.,z_{-i},\alpha,\alpha^*,\bm \pi,\bm \pi^a, \lambda, \left\{\theta_k\right\},\left\{\theta_k^a\right\},a_.,\gamma)
=P(z_i=k\vert x_.,z_{-i},\alpha,\alpha^*,\left\{\theta_k\right\},
\left\{\theta_k^a\right\},a_i)
\end{eqnarray*}
\begin{eqnarray*}
\propto \left\{ \begin{array}{lc}
P(z_i=k\vert z_{-i},\alpha,\theta_k)P(x_i\vert z_i=k,z_{-i},\theta_k,\alpha) & a_i=0 \\
P(z_i=k\vert z_{-i},\alpha^*,\theta_k^a)P(x_i\vert z_i=k,z_{-i},\theta_k^a,\alpha^*) & a_i=1 
\end{array}
\right.
\end{eqnarray*}
\begin{eqnarray}
=\left\{ \begin{array}{lc}
\frac{n_k}{(n+\alpha-1)}F(x_i\vert \theta_k) & a_i=0 \\
\frac{n_k}{(n+\alpha^*-1)}F(x_i\vert \theta_k^a) & a_i=1 
\end{array}
\right.\label{eq:incadz1}
\end{eqnarray}

where $\alpha^*=\frac{100}{1- p_i}$, $p_i$ is the probability of $x_i$ being anomalous and $K$ is the number of non-empty clusters. Thus, the posterior probability of forming a new cluster denoted by $K+1$ is given by:
\begin{eqnarray*}
P(z_i=K+1\vert x_.,z_{-i},\alpha,\alpha^*,\bm \pi,\bm \pi^a, \lambda, \left\{\theta_k\right\},\left\{\theta_k^a\right\},a_.,\gamma)
=P(z_i=K+1\vert x_i,z_{-i},\alpha,\alpha^*,\lambda,a_i)
\end{eqnarray*}
\begin{eqnarray*}
\propto \left\{ \begin{array}{lc}
P(z_i=K+1\vert z_{-i},\alpha,\lambda)P(x_i\vert z_i=K+1, z_{-i},\alpha,\lambda,a_i) & a_i=0 \\
P(z_i=K+1\vert z_{-i},\alpha^*,\lambda)P(x_i\vert z_i=K+1, z_{-i},\alpha^*,\lambda,a_i) & a_i=1 
\end{array}
\right.
\end{eqnarray*}
\begin{eqnarray}
=\left\{ \begin{array}{lc}
\frac{\alpha}{n+\alpha-1}\int F(x_i\vert \theta)G_0(\theta\vert\lambda)d\theta & a_i=0 \\
\frac{\alpha^*}{n+\alpha^*-1}\int F(x_i\vert \theta^a)G_0^{EV}(\theta^a\vert\lambda)d\theta^a & a_i=1
\end{array}
\right.\label{eq:incadz2}
\end{eqnarray}

Similarly, the parameters for clusters $k\in\{1,2,\dots,K\}$ are sampled from:
\begin{eqnarray}
\theta_k \propto G_0(\theta_k\vert \lambda)\mathcal{L}({\bm x_k}\vert \theta_k) & \text{if cluster is not anomalous}\label{eq:incadtehta1}\\
\theta_k^a \propto G_0^{EV}(\theta_k^a\vert \lambda)\mathcal{L}({\bm x_k}\vert \theta_k^a) & \text{if cluster is anomalous}\label{eq:incadtehta2}
\end{eqnarray}
 where $\bm x_k=\{x_i|z_i=k\}$ is the set of all points in cluster $k$. Finally, to identify the anomaly classification of the data, the posterior probability of $a_i$ is given by:
\begin{eqnarray*}
P(a_i=1\vert x_.,z_.,\alpha,\alpha^*,\bm \pi,\bm \pi^a, \lambda, \left\{\theta_k\right\},\left\{\theta_k^a\right\},\gamma)
=P(a_i=1\vert x_i,z_.,\alpha^*,\lambda, \left\{\theta_k^a\right\},\gamma)\\
\propto \sum_{k=1}^{K+1} P(a_i=1\vert x_i,z_i=k,z_{-i},\alpha^*,\lambda, \left\{\theta_k^a\right\},\gamma) *P(z_i=k\vert x_i,z_{-i},\alpha^*,\lambda, \left\{\theta_k^a\right\},\gamma)
\end{eqnarray*}

\begin{eqnarray}
=\sum_{k=1}^{K} P(x_i\vert \theta_k^a) \gamma \frac{n_k}{(n+\alpha^*-1)}
+\left(\int F(x_i\vert \theta^a)G_0^{EV}(\theta^a\vert\lambda)d\theta^a\right) \gamma \frac{\alpha^*}{n+\alpha^*-1}\label{eq:incada1}
\end{eqnarray}
Similarly,
\begin{eqnarray*}
P(a_i=0\vert x_i,z_.,\alpha,\lambda, \left\{\theta_k\right\},\gamma)
\end{eqnarray*}
\vspace{-7.5mm}
\begin{eqnarray}
\propto \sum_{k=1}^{K} P(x_i\vert \theta_k) (1-\gamma) \frac{n_k}{(n+\alpha-1)}
+\left(\int F(x_i\vert \theta)G_0(\theta\vert\lambda)d\theta\right) (1-\gamma) \frac{\alpha}{n+\alpha-1} \label{eq:incada2}
\end{eqnarray}

\begin{algorithm}[h]
\caption{Gibbs Sampling Algorithm when $G^{EV}_0$ is available}
\label{alg:generator1}
\SetKwProg{generate}{Function \emph{generate}}{}{end}
Given $z_.^{(t-1)},a_.^{(t-1)}, \left\{\theta_k^{(t-1)}\right\},\left\{\theta_k^{a(t-1)}\right\}$ from previous iterations. Let $K$ be the total number of clusters found till last iteration. Sample $z_.^{(t)},a_.^{(t)}, \left\{\theta_k^{(t)}\right\},\left\{\theta_k^{a(t)}\right\}$ as follows 
\begin{enumerate}
    \item Set $z_.=z_.^{(t-1)} $ and $a_.=a_.^{(t-1)}$
    \item \For{each point $i\rightarrow 1\ \KwTo\ N$}{
    \begin{enumerate}
        \item Remove $x_i$ from its cluster $z_i$.\label{step1}
        \item If $x_i$ is the only point in its cluster, $n_{z_i}=0$ after step (2)(a). Remove the cluster and \\update K to K-1.
        \item Rearrange cluster indices to ensure none are empty. 
        \item Sample $z_i$ from the Multinomial distribution given by Equations \ref{eq:incadz1} and \ref{eq:incadz2}\label{step2}
        \item If $z_i=K+1$, then sample new cluster parameters from the following distribution (It must be noted that the above posterior distribution was derived under the assumption of independence and exchangeability of priors for mathematical ease.)
\begin{eqnarray*}
\theta \left \vert x_i,z_., \left\{\theta_k^{(t-1)}\right\},
            \left\{\theta_k^{a(t-1)}\right\},a_.^{(t-1)} \right.
\end{eqnarray*}
\begin{eqnarray*}
            \propto \left\{ 
            \begin{array}{lc}
             \alpha G_0(\theta\vert\lambda)F(x_i\vert \theta)+ \sum_{j\neq i}F(x_i\vert \theta_{z_j})\delta(\theta-\theta_{z_j}^{(t-1)})\delta(a_j^{(t-1)})
             \quad ,a_i^{(t-1)}=0\\
             \alpha^* G_0^{EV}(\theta\vert\lambda)F(x_i\vert \theta)+
            \sum_{j\neq i} F(x_i\vert \theta_{z_j})\delta(\theta-\theta_{z_j}^{(t-1)})\delta(a_j^{(t-1)}-1)
             \quad ,a_i^{(t-1)}=1
            \end{array}
            \right.
\end{eqnarray*} 
Update K=K+1.
   
    \item For each cluster $k\in \{1,2,\dots,K\}$, sample cluster parameters $\theta_k$ and $\theta_k^a$ using \\Equations \ref{eq:incadtehta1} and \ref{eq:incadtehta2}.

    \item Sample the anomaly classification $a_i$ from the Binomial distribution given by \\Equations \ref{eq:incada1} and \ref{eq:incada2}.

    \item Set $z_.^{(t)}=z.$ and $a_.^{(t)}=a_.$
     \end{enumerate}
    }
\end{enumerate}
\end{algorithm}

\subsection{Inference when $G^{EV}_0$ is not available}
The estimation of $G^{EV}_0$ is required on two occasions. Firstly, while sampling the parameters of anomalous clusters when generating the data and estimating the posterior distribution. Secondly, to compute the probability of the point being anomalous when estimating the updated concentration parameter $\alpha_2$. When estimating $G_0^{EV}$ is not feasible, the following two modifications to the original model are proposed:
\begin{enumerate}
    \item Since an approximate $G_0^{EV}$ distribution need not belong to the family of conjugate priors of $F$, we need a different approach to sample the parameters for anomalous clusters. Thus, we assume $\theta^a\sim G_0$ for sampling the parameters $\{\theta^a_k\}_{k=1}^\infty$ for anomalous clusters.
    \item To estimate the probability of a point being anomalous, use the approach described by \citet{Clifton:2014}.
\end{enumerate}
The pseudo-Gibbs sampling algorithm, presented in Algorithm \ref{alg:generator2}, has been designed to address the cases when $G^{EV}_0$ is not available. For such cases, the modified $f$ is given by, 
\begin{eqnarray*}
    f(\alpha|x_n,\textbf{x},\textbf{z})=
    \left\{\begin{array}{cc}
         \alpha &, if\ not\ in\ tail\\
         \alpha*(1-ev\_prop)+\frac{100}{1-p_n}*ev\_prop &,if\ in\ tail\\
    \end{array}\right.
\end{eqnarray*}
where $ev\_prop$ determines the effect of a anomalous behavior on the concentration parameter\footnote{It can be seen that when the data point $x_n$ has extreme or rare features differentiating it from all existing clusters, the corresponding density $f_y(x_n)$ decreases. This makes it a left tail point in the distribution $G_y$. The farther away it is in the tail, the lower the probability of its $\delta$-nbd being in the tail and hence a higher $f(\alpha|x_n,\textbf{x},\textbf{z})$.}. $ev\_prop$ is solely used to speed up the convergence in the Gibbs sampling. Since the estimation of the $G^{EV}$ distribution is not always possible, an alternate/pseudo Gibbs sampling algorithm has been presented in Algorithm \ref{alg:generator2}. 

\begin{algorithm}
\caption{Gibbs Sampling Algorithm when $G^{EV}_0$ is not available}
\label{alg:generator2}
\SetKwProg{generate}{Function \emph{generate}}{}{end}
Given $z_.^{(t-1)},a_.^{(t-1)}, \left\{\theta_k^{(t-1)}\right\},\left\{\theta_k^{a(t-1)}\right\}$ from previous iterations. Let $K$ be the total number of clusters found till last iteration. Sample $z_.^{(t)},a_.^{(t)}, \left\{\theta_k^{(t)}\right\},\left\{\theta_k^{a(t)}\right\}$ as follows 
\begin{enumerate}
    \item Set $z_.=z_.^{(t-1)} $ and $a_.=a_.^{(t-1)}$
    \item \For{each point $i\rightarrow 1\ \KwTo\ N$}{
    \begin{enumerate}
    \item Steps \ref{step1} $\KwTo$ \ref{step2} in Algorithm \ref{alg:generator1}  \label{step3}
        \item If $z_i=K+1$, then set the cluster distribution to be multivariate normal with the new \\cluster mean as $x_i$ and cluster variance as $\Sigma$ which is pre-defined.

Update K=K+1.

    \item For each cluster $k\in \{1,2,\dots,K\}$, sample cluster parameters $\theta_k$ and $\theta_k^a$ using \\Equation \ref{eq:incadtehta1}.

    \item Sample the anomaly classification $a_i$ from the Binomial($p_i$) where $p_i$ is given by \\Equation \ref{eq:incadpi}. If most of the cluster instances are classified as anomalous, classify all of \\the cluster's instances as anomalies. \label{step4}

    \item Set $z_.^{(t)}=z.$ and $a_.^{(t)}=a_.$ 
     \end{enumerate}
    }
\end{enumerate}
\end{algorithm}
\subsection{Exchangeability}
A model is said to be exchangeable when for any permutation $S$ of $\{1,2,...,n\}$,
$P(x_1,x_2,...x_n)=P(x_{S(1)},x_{S(2)},...x_{S(n)})$.
Looking at the joint probability of the cluster assignments for the integrated model, we know,
\begin{eqnarray*}
    P(z_1,z_2,..z_n|\textbf{x})=P(z_1|\textbf{x})P(z_2|z_1,\textbf{x})..P(z_n|z_{1:n-1},\textbf{x})
\end{eqnarray*}

Without loss of generality, let us assume there are $K$ clusters. Let, for any $k<K$, the joint probability of all the points in cluster $k$ be given by

\begin{eqnarray*}
    \left( \frac{\alpha*p_{k,1}}{I_{k,1}+\alpha-1}+\frac{\alpha^**(1-p_{k,1})}{I_{k,1}+\alpha^* -1}\right)
    \prod_{n_k=2}^{N_k}\left( \frac{(n_k-1)*p_{k,n_k}}{I_{k,n_k}+\alpha-1}+\frac{(n_k-1)*(1-p_{k,n_k})}{I_{k,n_k}+\alpha^* -1}\right)\\
\end{eqnarray*}

where $N_k$ is the size of the cluster $k$ , $I_{k,i}$ is the index of the $i^{th}$ instance joining the $k^{th}$ cluster and $p_{k,i}=p_{I_{k,i}}$. Thus, the joint probability for complete data is then given by
\begin{eqnarray*}
    \frac{\prod_{k=1}^{K}\left[(I_{k,1}-1)p_{k,1}(\alpha-\alpha^*)+\alpha^*(I_{k,1}+\alpha-1)
    \prod_{n_k=2}^{N_k}(n_k-1)(I_{k,n_k}+\alpha-1+p_{k,n_k}(\alpha^*-\alpha)
    \right]}{\prod_{i=1}^{N}\left((i+\alpha-1)(i+\alpha^*-1) 
    \right)}
\end{eqnarray*}

which is dependent on the order of the data. This shows that the model is not exchangeable unless $\alpha=\alpha^*$ or $p_{k,n_k}=0$ or $p_{k,n_k}=1$. These conditions effectively reduce the prior distribution to a traditional CRP model. Hence, it can be  concluded that the INCAD model cannot be modified to be exchangeable.

\subsubsection{Non-exchangeable models in streaming settings}
Though exchangeability is a reasonable assumption in many situations, the evolution of behavior over time is not captured by traditional exchangeable models. In particular for streaming settings, using non-exchangeable models captures  the effect of the order of the data. In such settings, instances that are a result of new evolving behavior should be monitored (as anomalous) until the behavior becomes relatively prevalent. Similarly, relapse of outdated behaviors (either normal or anomalous) should also be subjected to critical evaluation due to extended lag observed between similar instances. Such order driven dependency can be well captured in non-exchangeable models making them ideal for studying streaming data. 
\subsection{Adaptability to sequential data}
One of the best outcomes of having a non-exchangeable prior is its ability to capture the {\em drift or evolution} in the behavior(s) either locally or globally or a mixture of both. INCAD model serves as a perfect platform to detect these changes and delivers an adaptable classification and clustering. The model has a straightforward extension to sequential settings where the model evolves with every incoming instance. Rather than updating the model for entire data with each new update, the streaming INCAD model re-evaluates only the tail instances. This enables the model to identify the following evaluations in the data:
\begin{enumerate}
    \item New trends that are classified as anomalous but can eventually grow to become normal.
    \item Previously normal behaviors that have vanished over time but have relapsed and hence become anomalous (eg. disease relapse post complete recovery)
\end{enumerate}
The Gibbs sampling algorithm for the streaming INCAD model is given in Algorithm \ref{alg:generator3}. 
\begin{figure}
\centering
    \includegraphics[width=3.5cm]{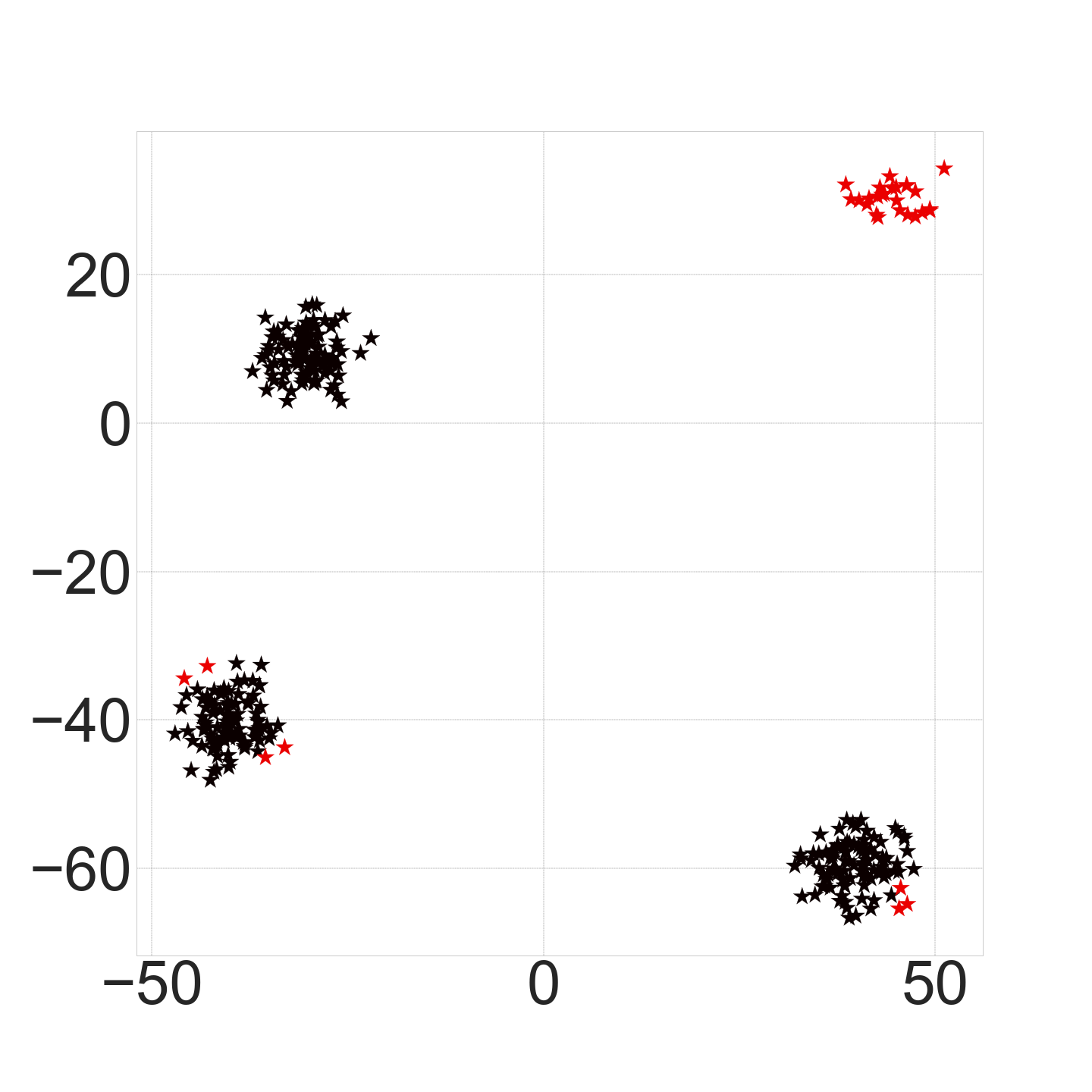}
    \label{fig:fig00} 
    \includegraphics[width=3.5cm]{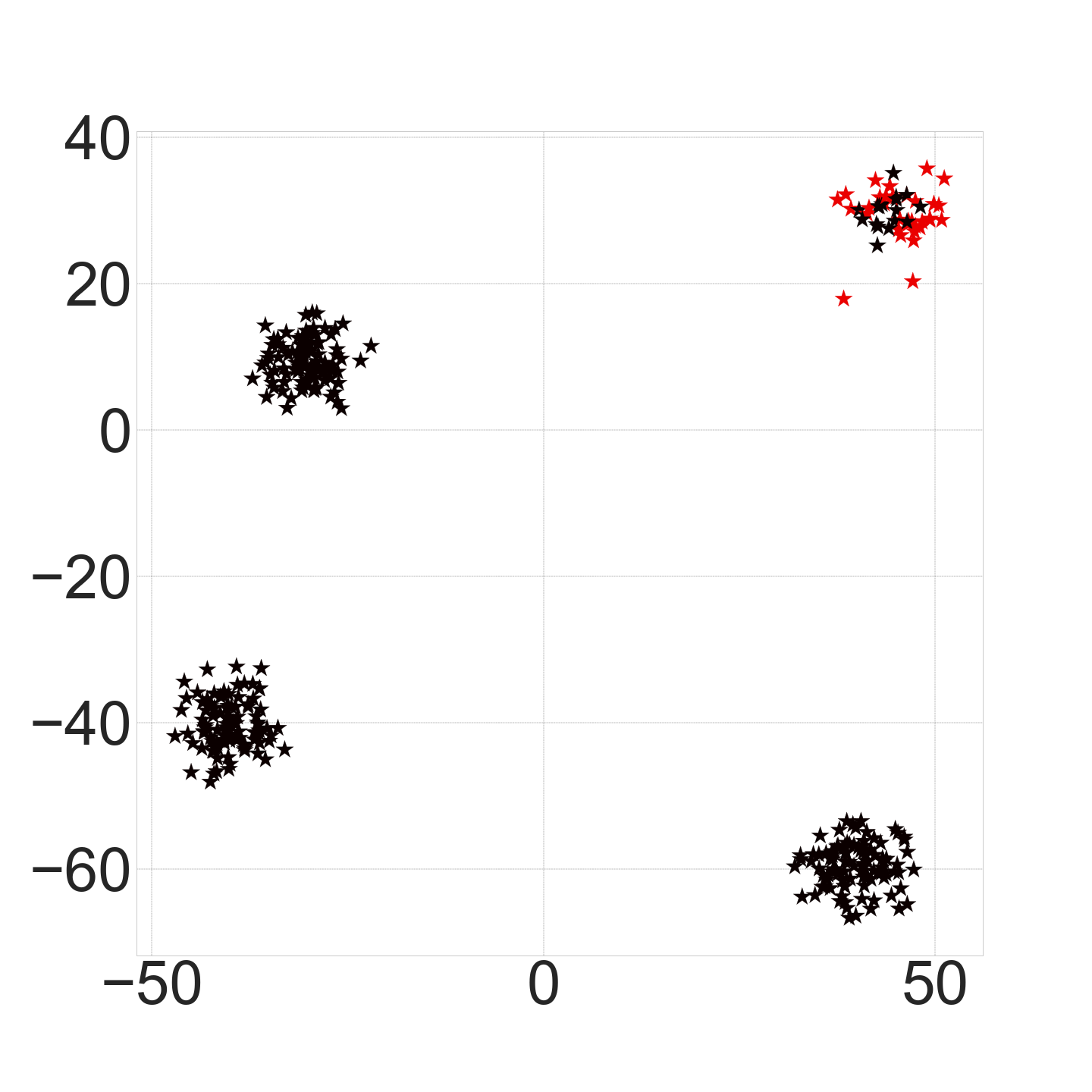}
    \label{fig:fig01}
    \includegraphics[width=3.5cm]{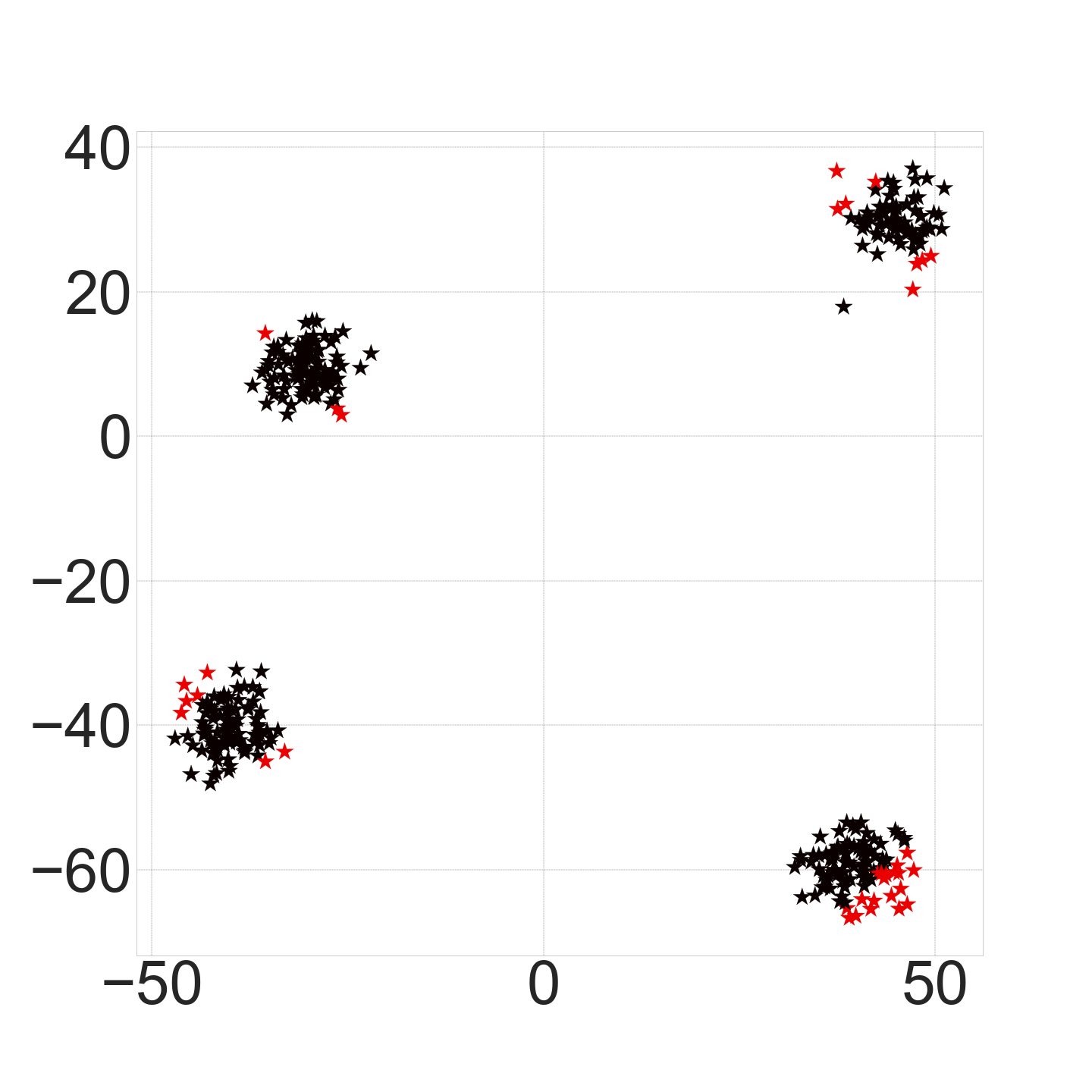}
    \label{fig:fig02}
    \includegraphics[width=3.5cm]{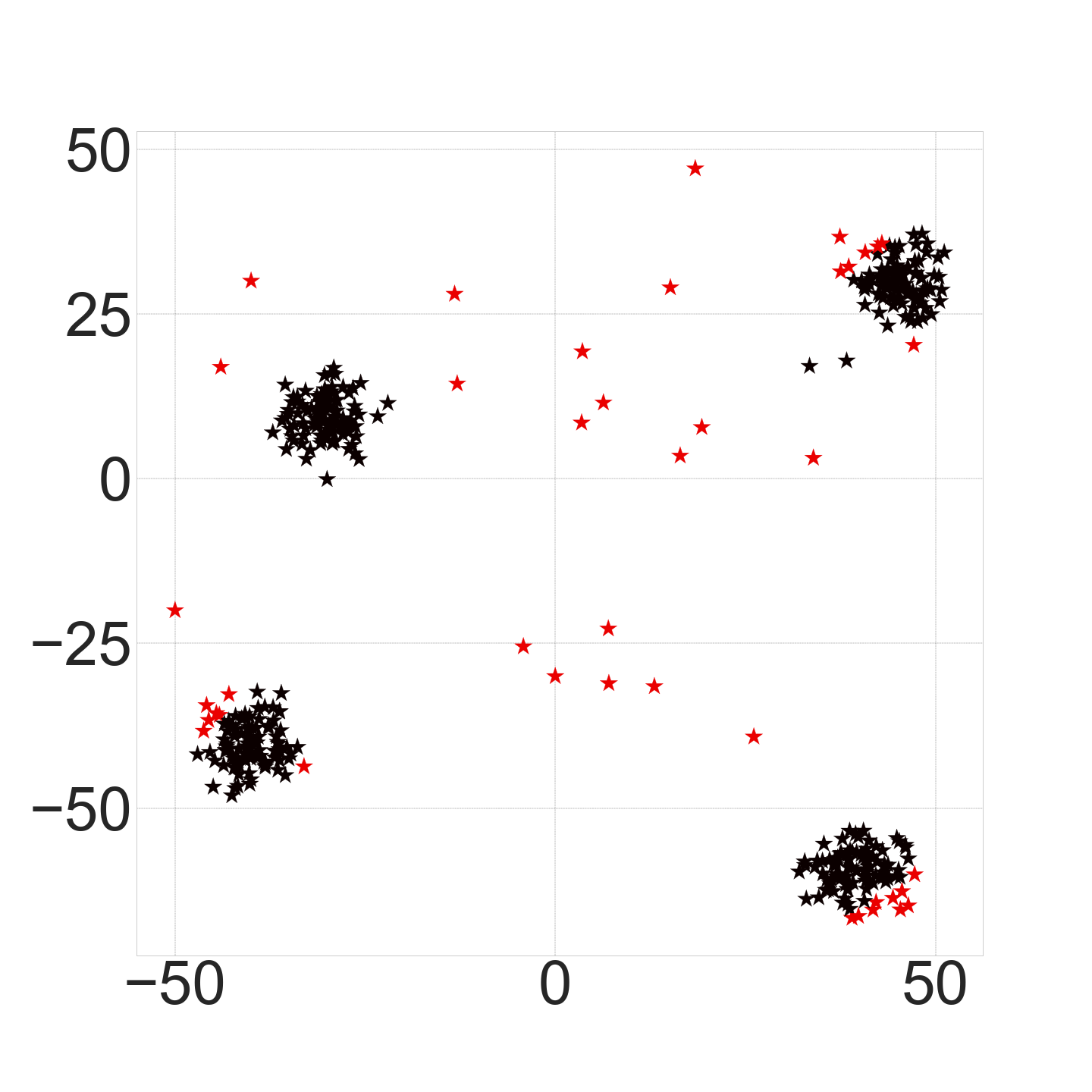}
    \caption{Evolution of anomaly classification using streaming INCAD: The classification into anomalous and normal instances are represented by {\rdot} and {\bdot} respectively. Note the evolution of the classification of top right cluster from anomalous to normal with incoming data}
    \label{fig:fig03}
\end{figure}
\begin{figure}
\centering
    \includegraphics[width=3.5cm]{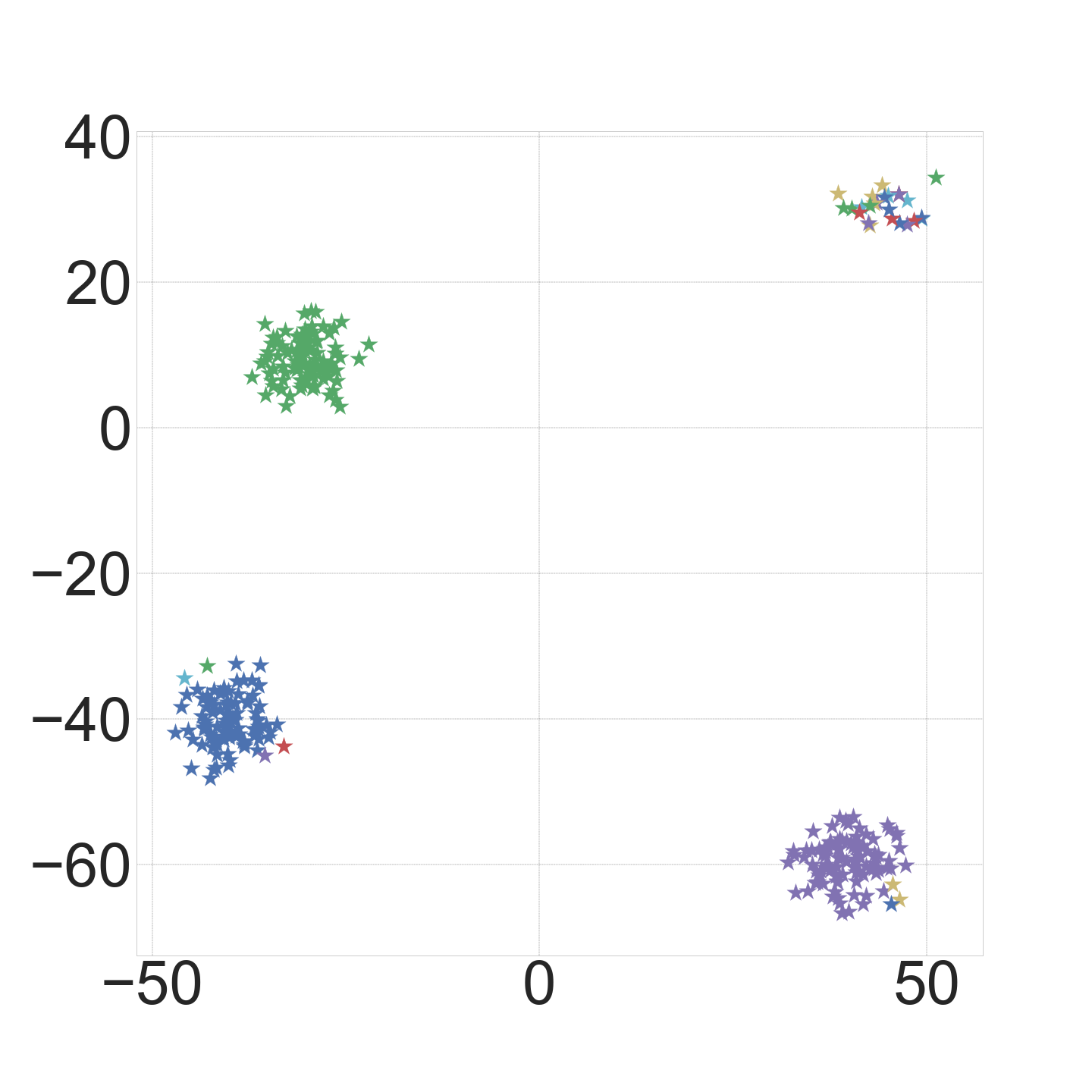}
    \label{fig:fig10}
    \includegraphics[width=3.5cm]{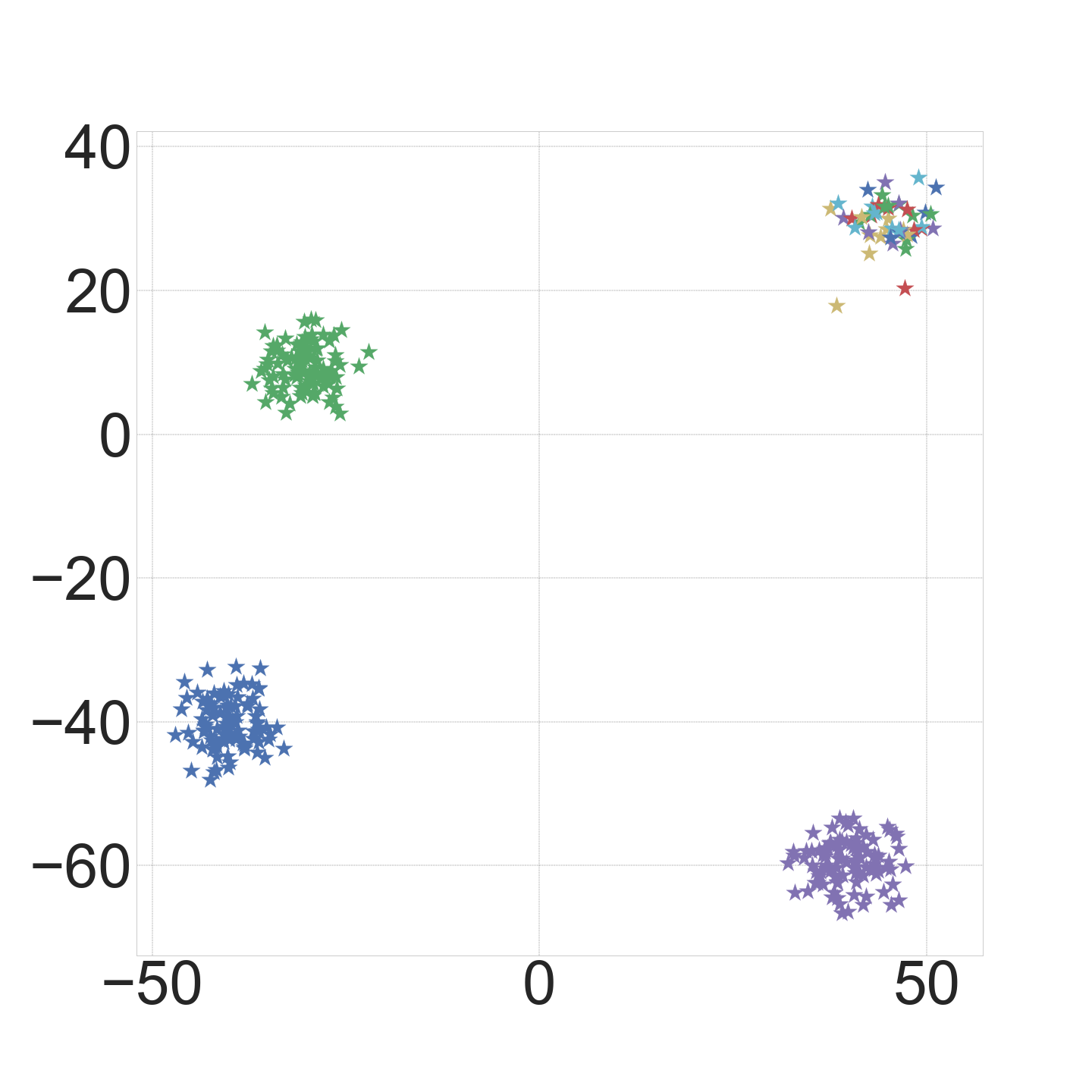}
    \label{fig:fig11}
    \includegraphics[width=3.5cm]{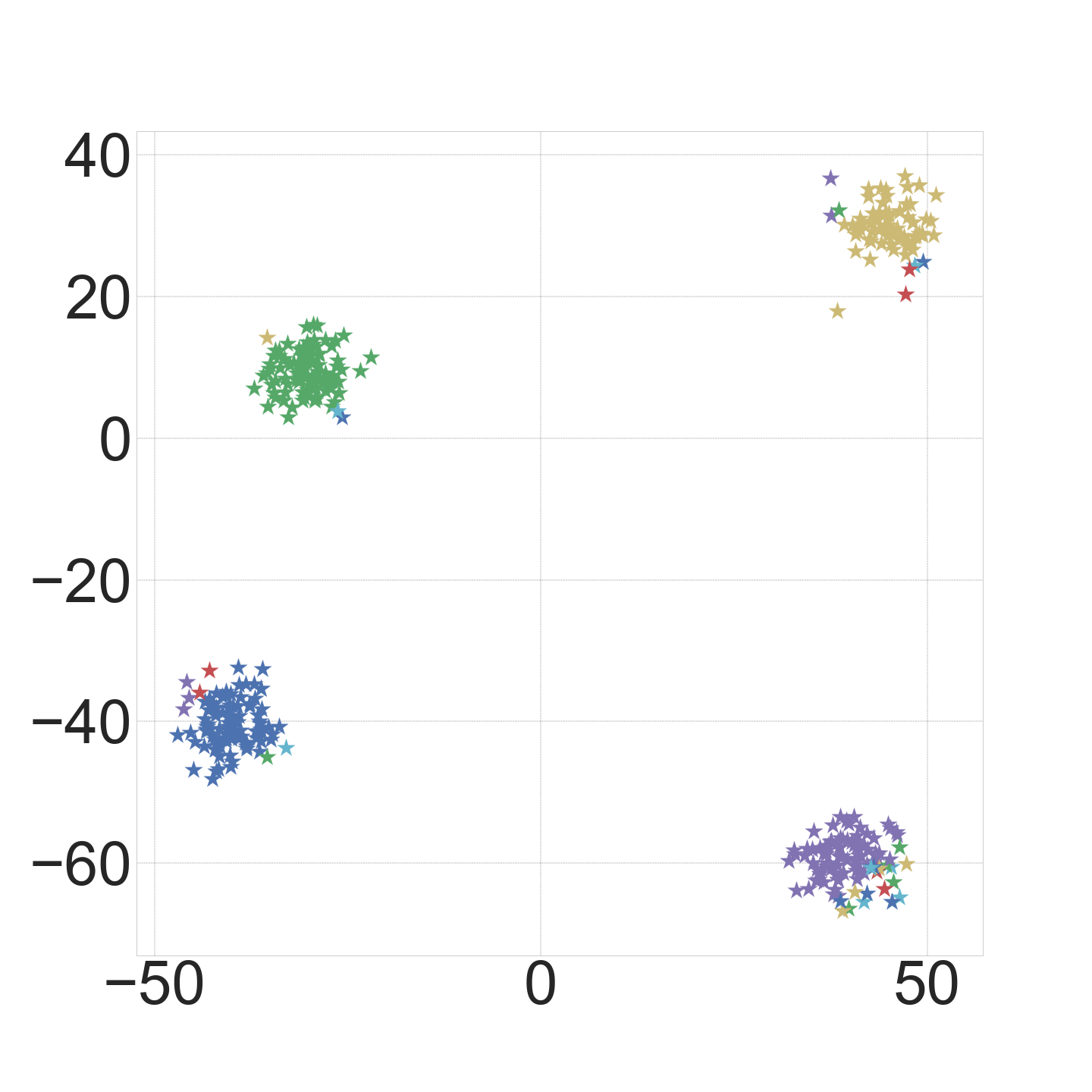}
    \label{fig:fig111}
    \includegraphics[width=3.5cm]{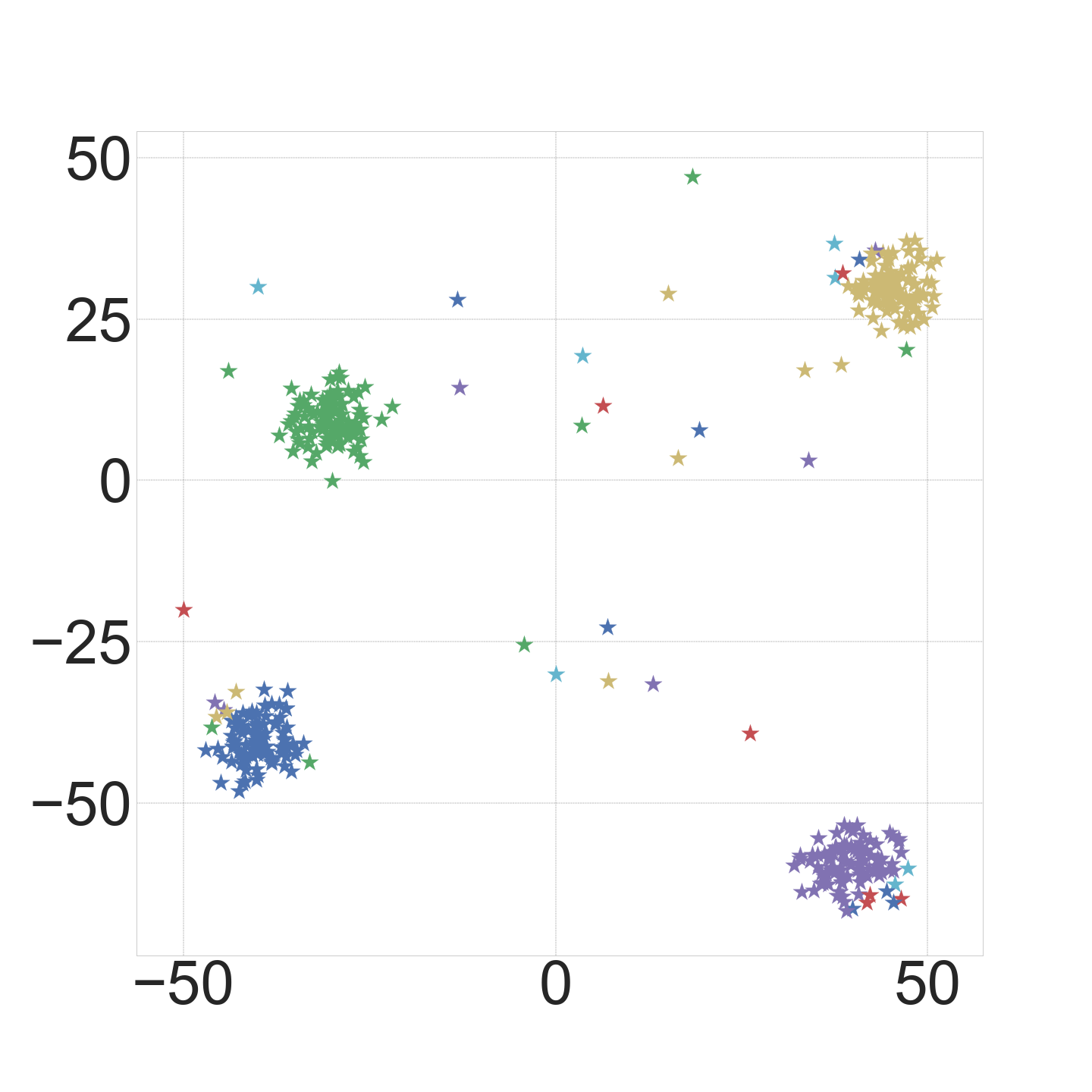}
    \caption{Evolution of clustering using streaming INCAD: Each cluster is denoted by a different color. Notice the evolution of random points in top right corner into a well formed cluster in the presence of more data}
    \label{fig:fig13}
\end{figure}

\begin{algorithm}
\caption{Algorithm for Streaming Extension}
\label{alg:generator3}
Perform clustering on a small portion of the data (10-20\%) using non-streaming model\\
Set $ev_{prop}=exp^{-0.5}$\\
\For{each new data point $x_N$}{
\begin{enumerate}
        \item Compute the mixture proportions $m\_para$ and the mixture density for all the data. 
        
        Compute ${t_1} = q^{th}$ percentile pdf value to identify the tail points
        \item For each $x_i\ s.t.\ f(x_i)<t_1$ repeat steps \ref{step3}$\to $\ref{step4} of Algorithm \ref{alg:generator2}
        \end{enumerate}
}
If cluster size $ \leq 0.05*N$ then, classify all the cluster points as anomalies.
\end{algorithm}
%

\section{Results}\label{sec5}
In this section, we evaluate the proposed model using benchmark streaming datasets from NUMENTA. The streaming INCAD model's anomaly detection is compared with SPOT algorithm developed by \citet{Siffer:2017}. The evolution of clustering and anomaly classification using the streaming INCAD model is visualized using simulated dataset. In addition, the effect of batch vs stream proportion on quality of performance is presented. For Gibbs sampling initialization, the data was assumed to follow a mixture of MVN distributions. 10 clusters were initially assumed with the same initial parameters. The cluster means were set to the sample mean and the covariance matrix as a multiple of the sample covariance matrix, using a scalar constant. The concentration parameter $\alpha$ was always set to 1. 

\subsection{Simulated Data}\label{simulated}
For visualizing the model's clustering and anomaly detection, a 2-dimensional data set of size 400 with 4 normal clusters and 23 anomalies sampled from a normal distribution centered at (0,0) and a large variance was generated for model evaluation. Small clusters and data outliers are regarded as true anomalies. Data from the first 3 normal clusters (300 data points) were first modeled using non-streaming INCAD. The final cluster and the anomalies were then used as updates for the streaming INCAD model. The evolution in the anomaly classification is presented in Figure \ref{fig:fig03}.
\begin{figure}
    \centering
    \includegraphics[width=4.75cm]{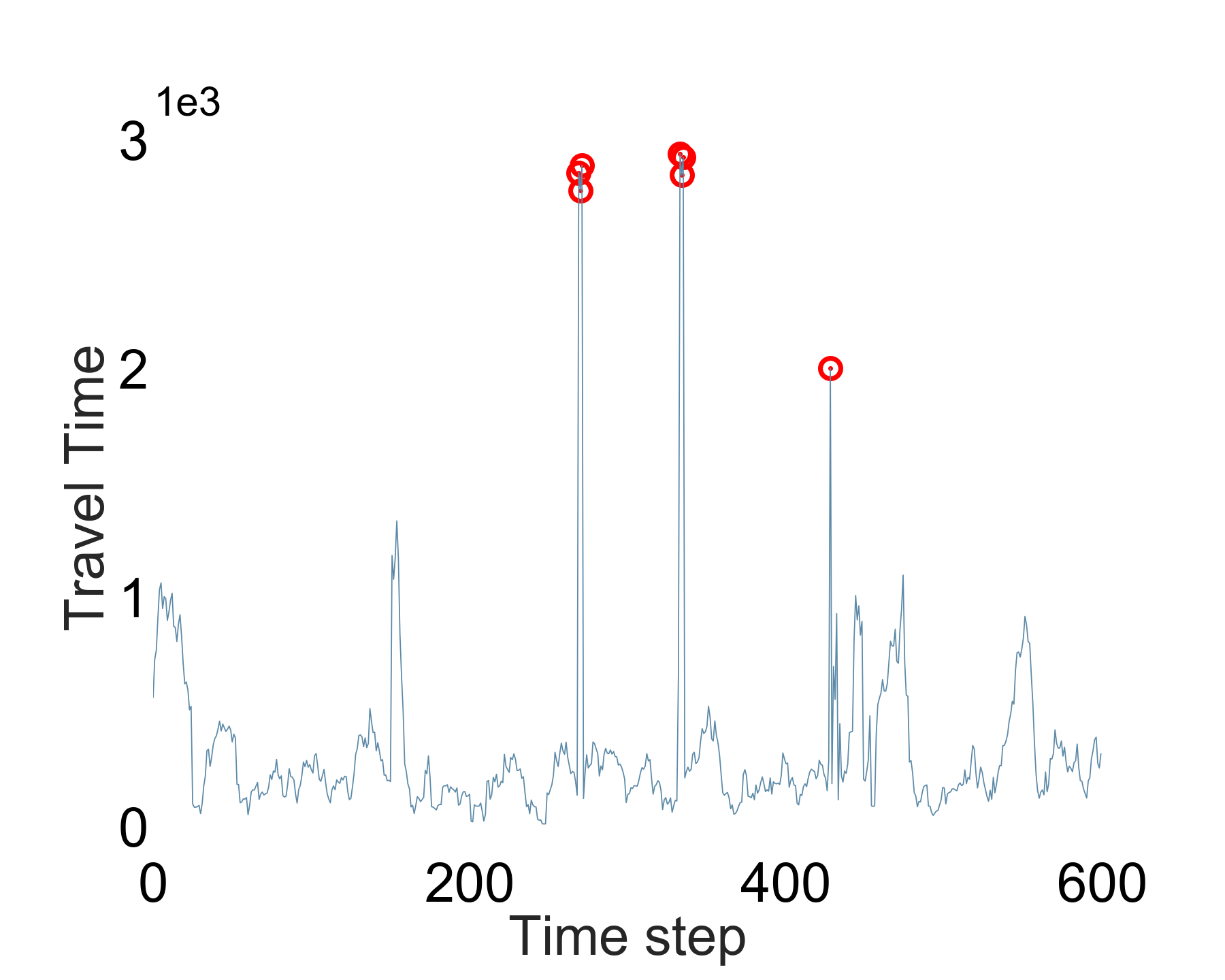}
    \centering
    \includegraphics[width=4.75cm]{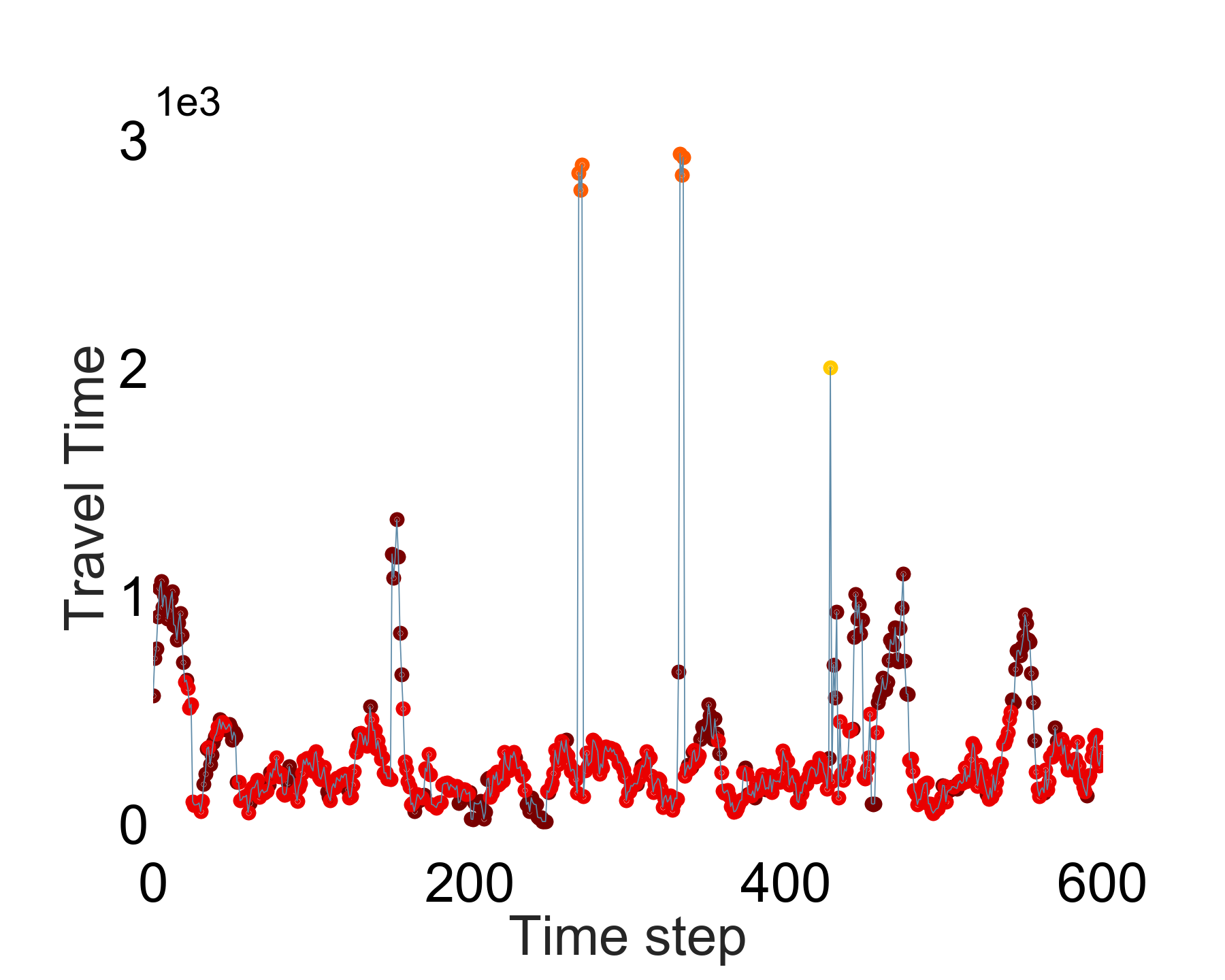}
    \centering
    \includegraphics[width=4.75cm]{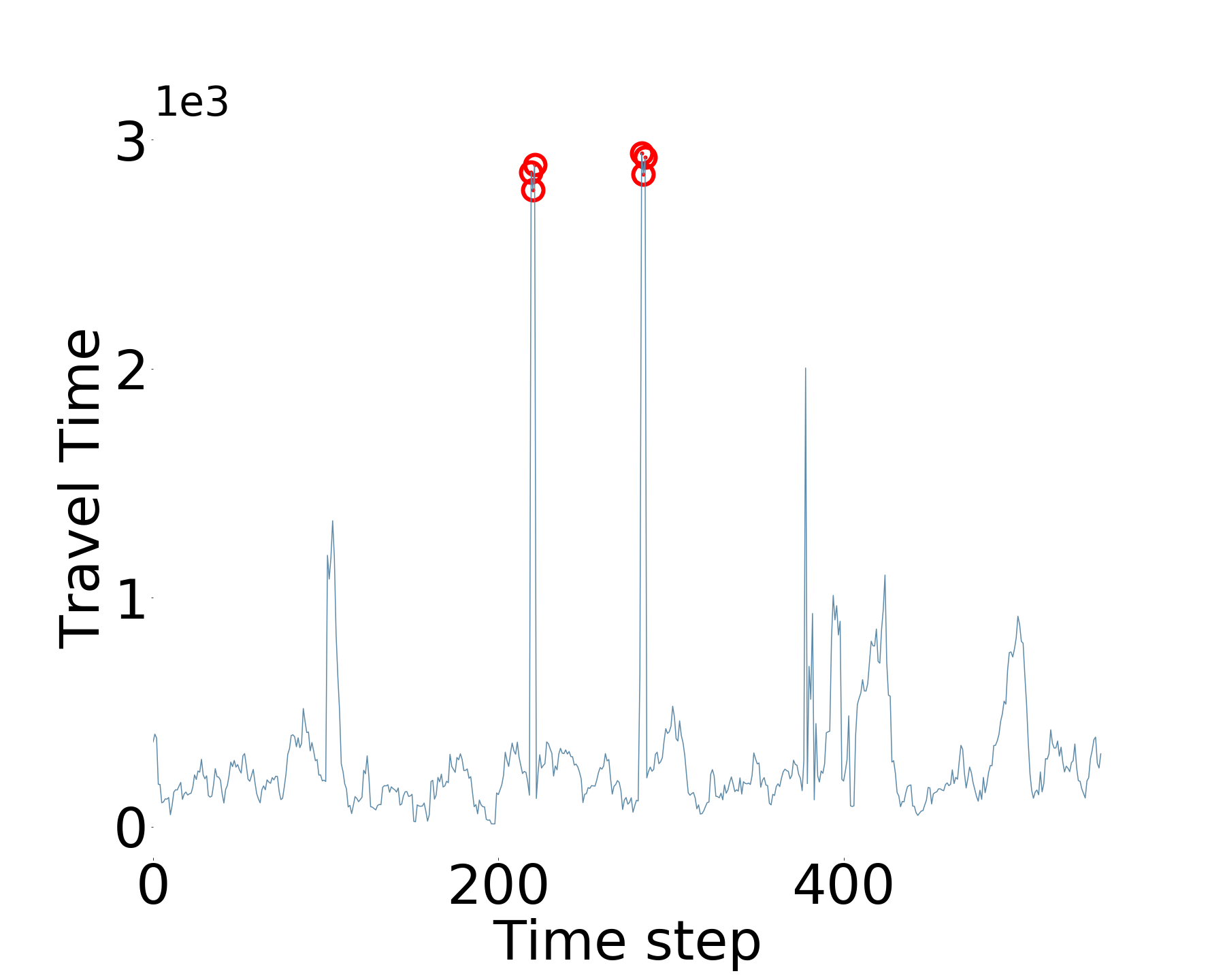}
    \caption{NUMENTA traffic data: (from left to right) Anomaly detection and clustering using streaming INCAD and anomaly detection using SPOT \cite{Siffer:2017}}
    \label{fig:fig11}
\end{figure}
\begin{figure}
    \centering
    \includegraphics[width=4.75cm]{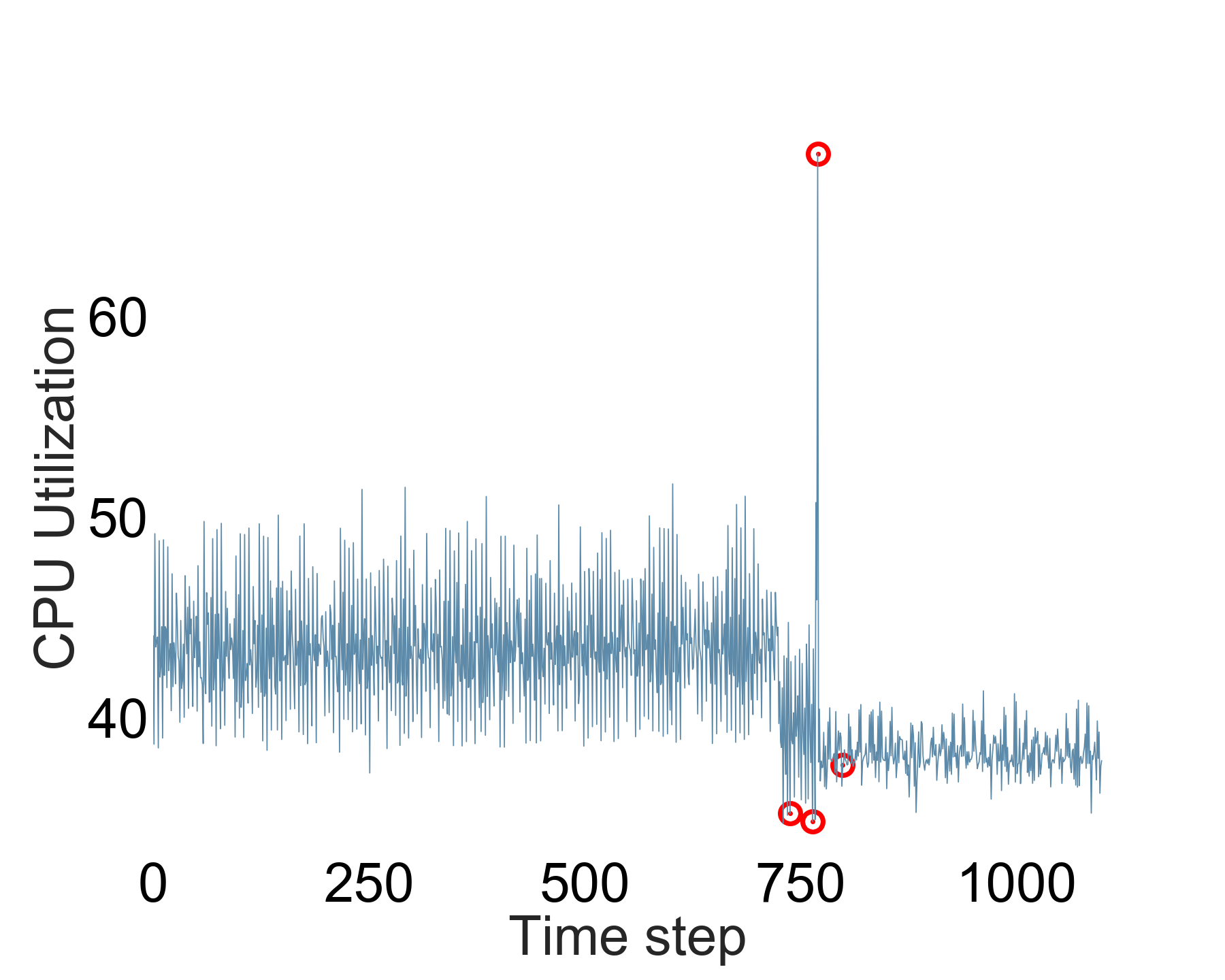}
    \centering
    \includegraphics[width=4.75cm]{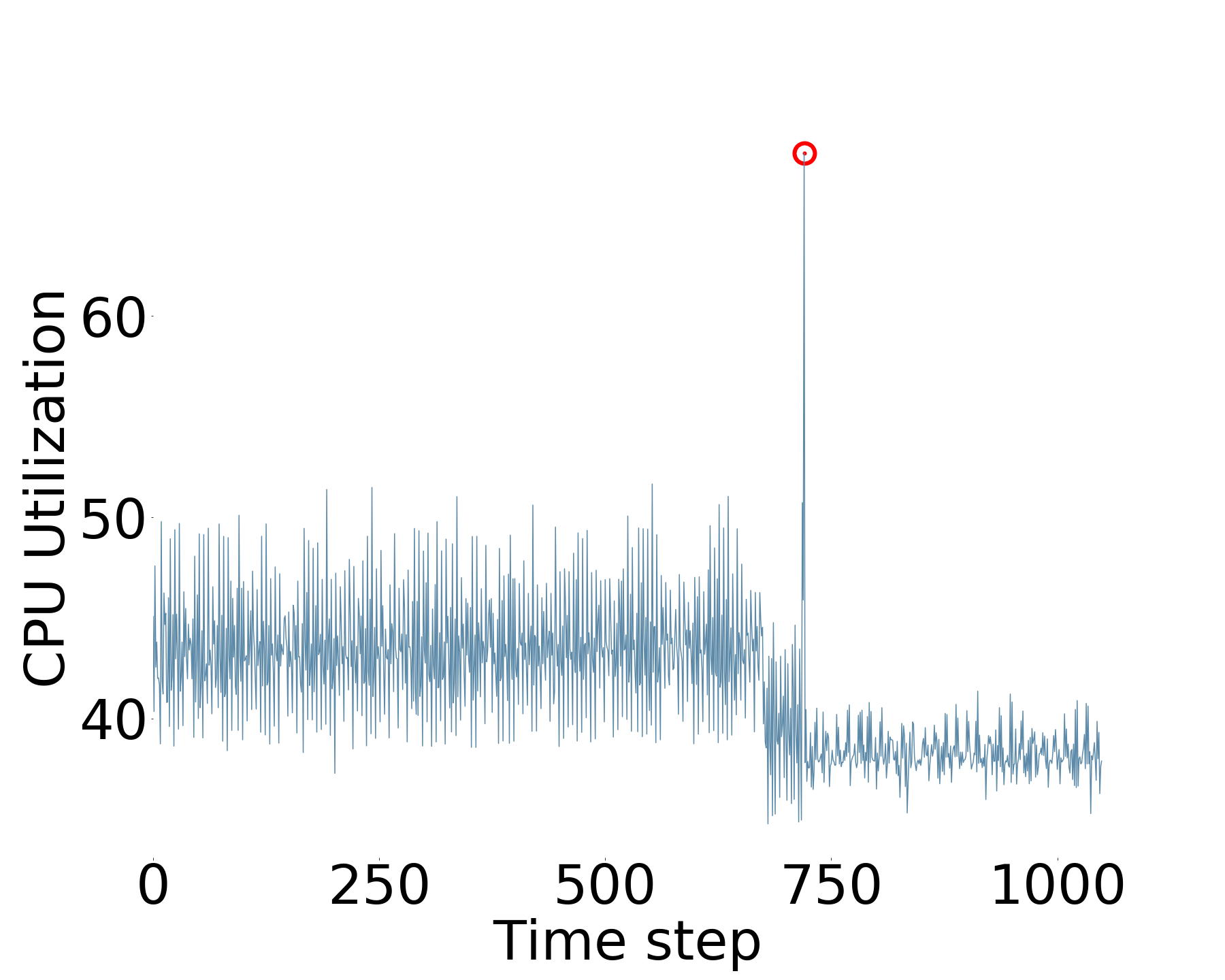}
    \caption{Anomaly detection on NUMENTA AWS cloud watch data using streaming INCAD(left) and SPOT \cite{Siffer:2017} (right)}
    \label{fig:fig14}
\end{figure}
\subsection{Model Evaluation on NUMENTA Data}
Two data sets from NUMENTA \cite{numentapressrelease} namely the real traffic data and the AWS cloud watch data were used as benchmarks for the streaming anomaly detection. The streaming INCAD model was compared with the SPOT algorithm developed by \citet{Siffer:2017}. Unlike SPOT algorithm, the streaming INCAD is capable of modeling data with more than one feature. Thus, the data instance, as well as the time of the instance, were used to develop the anomaly detection model. Since the true anomaly labels are not available, the model's performance with respect to SPOT algorithm was evaluated based on the ability to identify erratic behaviors. The model results on the datasets using streaming INCAD and SPOT have been presented in Figures \ref{fig:fig11} and \ref{fig:fig14}. 
\begin{figure}
    \centering
    \includegraphics[width=4.25cm]{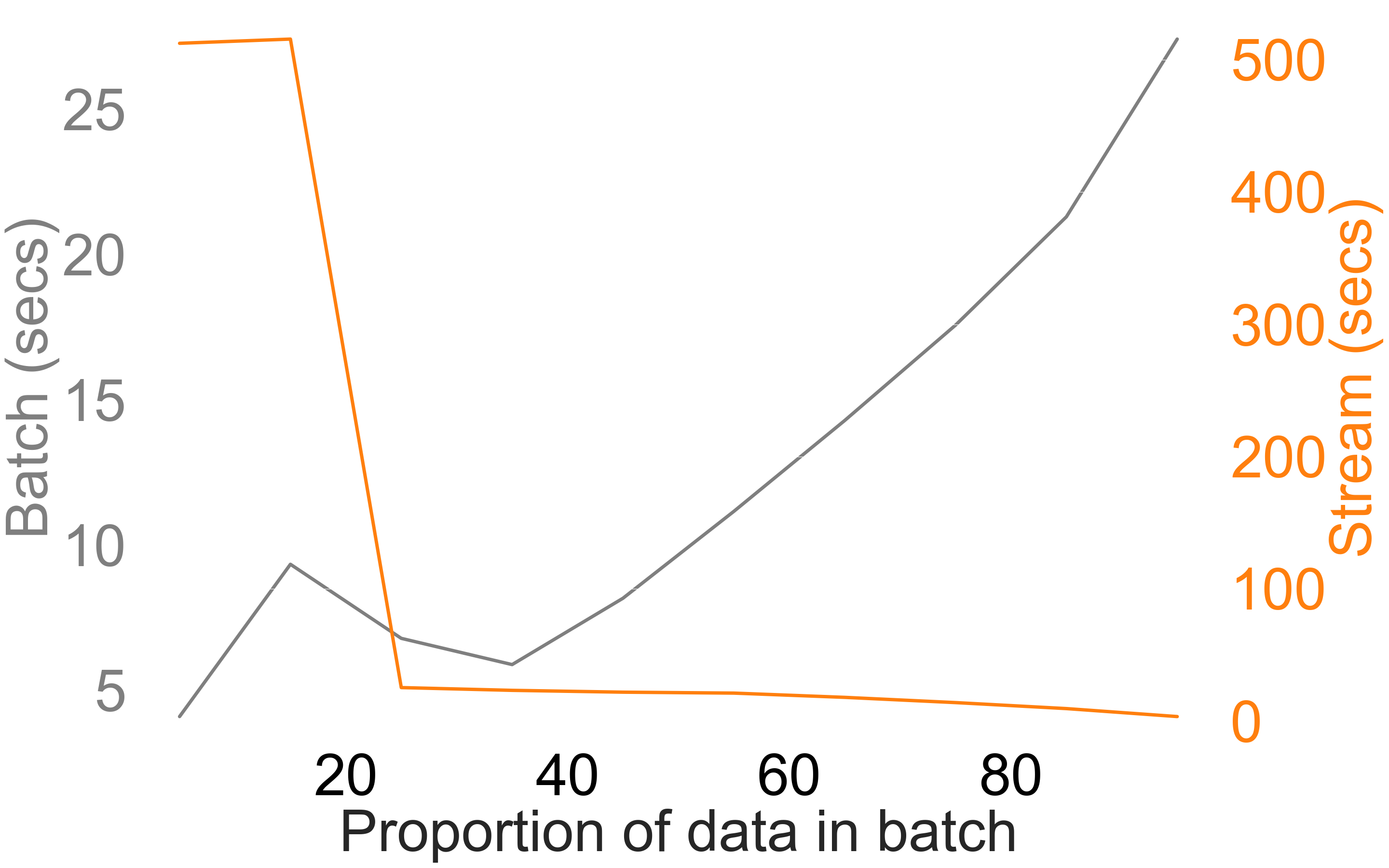}
    \centering
    \includegraphics[width=4.75cm]{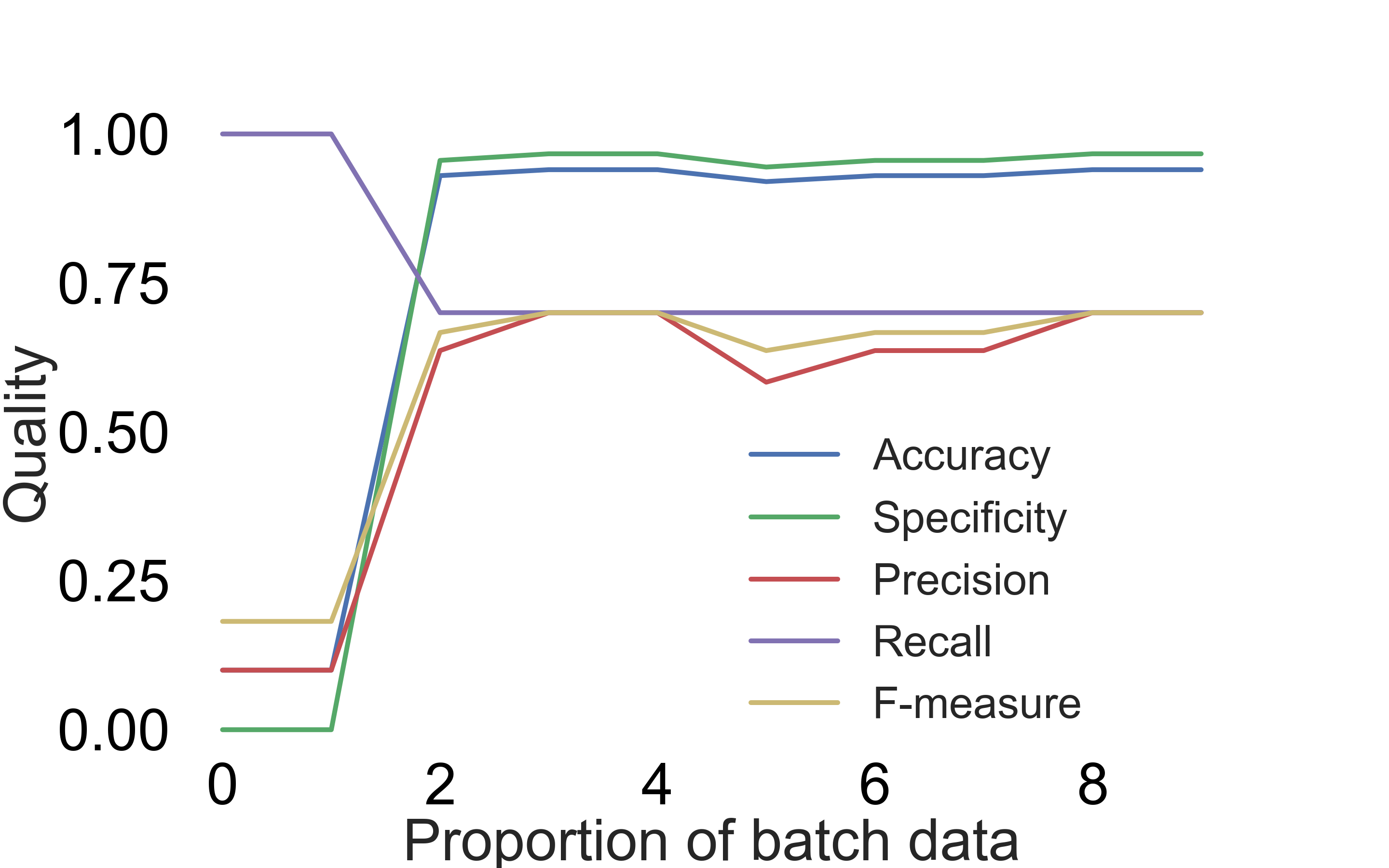}

    \caption{Proportion of data for batch model vs Quality: Computational time (left) and Proportion of data for the batch model (in tens) vs Quality of Anomaly detection (right) }
    \label{fig:fig8}
\end{figure}
\vspace{-10mm}
\section{Sensitivity to Batch Proportion}
Streaming INCAD model re-evaluates the tail data at each update, the dependency of the model's performance on the current state must be evaluated. Thus, various metrics were used to study the model's sensitivity to the initial batch proportion. Figure \ref{fig:fig8} shows the effect of batch proportion on computational time and performance of anomaly detection. The simulated data defined in Section \ref{simulated} was used for the sensitivity analysis. It can be seen that the computational time is optimal for $~25\%$ of data used in to run the non-streaming INCAD model. As anticipated, precision, accuracy, specificity, and f-measure for the anomaly detection were observed to plateau after a significant increase.

\section{Conclusion and Future Work}\label{sec6}
A detailed description of the INCAD algorithm and the motivation behind it has been presented in this paper. The model's definition of an anomaly and its adaptable interpretation sets the model apart from the rest of the clustering based anomaly detection algorithms. While past anomaly detection methods lack the ability to simultaneously perform clustering and anomaly detection or to the INCAD model not only defines a new standard for such integrated methods but also breaks into the domain of streaming anomaly detection. The model's ability to identify anomalies and cluster data using a completely data-driven strategy permits it to capture the evolution of multiple behaviors and patterns within the data. 

Additionally, the INCAD model can be smoothly transformed into a streaming setting. The model is seen to be robust to the initial proportion of the data subset that was evaluated using the non-streaming INCAD model. Moreover, this sets up the model to be extended to distribution families beyond multivariate normal. Though one of the key shortcomings of the model is its computational complexity in Gibbs sampling in the DPMM clusters, the use of faster methods such as variational inference might prove to be useful. 

\section{Acknowledgements}
The authors would like to acknowledge University at Buffalo Center for Computational Research (http://www.buffalo.edu/ccr.html) for its computing resources that were made available for conducting the research reported in this paper. Financial support of the National Science Foundation Grant numbers NSF/OAC 1339765 and NSF/DMS 1621853 is acknowledged.

\bibliographystyle{plainnat}
\bibliography{bibliography}

\begin{thebibliography}{29}
\providecommand{\natexlab}[1]{#1}
\providecommand{\url}[1]{\texttt{#1}}
\expandafter\ifx\csname urlstyle\endcsname\relax
  \providecommand{\doi}[1]{doi: #1}\else
  \providecommand{\doi}{doi: \begingroup \urlstyle{rm}\Url}\fi

\bibitem[Al-Behadili et~al.(2016)Al-Behadili, Grumpe, Migdadi, and
  W{\"o}hler]{al2016semi}
Husam Al-Behadili, Arne Grumpe, Lubaba Migdadi, and Christian W{\"o}hler.
\newblock Semi-supervised learning using incremental support vector machine and
  extreme value theory in gesture data.
\newblock In \emph{Computer Modelling and Simulation (UKSim)}, pages 184--189.
  IEEE, 2016.

\bibitem[Amer and Goldstein(2012)]{amer2012nearest}
Mennatallah Amer and Markus Goldstein.
\newblock Nearest-neighbor and clustering based anomaly detection algorithms
  for rapidminer.
\newblock In \emph{Proc. of the 3rd RCOMM 2012}, pages 1--12, 2012.

\bibitem[Antoniak(1974)]{Antoniak:1974}
Charles~E. Antoniak.
\newblock Mixtures of dirichlet processes with applications to bayesian
  nonparametric problems.
\newblock \emph{Annals of Statistics}, 2\penalty0 (6), 1974.

\bibitem[Bay and Schwabacher(2003)]{bay2003mining}
Stephen~D Bay and Mark Schwabacher.
\newblock Mining distance-based outliers in near linear time with randomization
  and a simple pruning rule.
\newblock In \emph{Proc. of the 9th ACM SIGKDD international conference on
  Knowledge discovery and data mining}, pages 29--38. ACM, 2003.

\bibitem[Blei and Frazier(2010)]{blei2010distance}
David~M Blei and Peter~I Frazier.
\newblock Distance dependent chinese restaurant processes.
\newblock In \emph{ICML}, pages 87--94, 2010.

\bibitem[Blei and Jordan(2004)]{Blei:2004}
David~M. Blei and Michael~I. Jordan.
\newblock Variational methods for the dirichlet process.
\newblock In \emph{Proc. of the Twenty-first International Conference on
  Machine Learning}, pages 12--, 2004.

\bibitem[Chan et~al.(2003)Chan, Mahoney, and Arshad]{chan2003machine}
Philip~K Chan, Matthew~V Mahoney, and Muhammad~H Arshad.
\newblock A machine learning approach to anomaly detection.
\newblock Technical report, 2003.

\bibitem[Chandola et~al.(2009)Chandola, Banerjee, and
  Kumar]{chandola2009anomaly}
Varun Chandola, Arindam Banerjee, and Vipin Kumar.
\newblock Anomaly detection: A survey.
\newblock \emph{ACM computing surveys (CSUR)}, 41\penalty0 (3):\penalty0 15,
  2009.

\bibitem[Clifton et~al.(2014)Clifton, Clifton, Hugueny, and
  Tarassenko]{Clifton:2014}
David~A Clifton, Lei Clifton, Samuel Hugueny, and Lionel Tarassenko.
\newblock Extending the generalised pareto distribution for novelty detection
  in high-dimensional spaces.
\newblock \emph{Journal of Signal Processing Systems}, 74\penalty0
  (3):\penalty0 323--339, 2014.

\bibitem[Eskin et~al.(2002)Eskin, Arnold, Prerau, Portnoy, and
  Stolfo]{eskin2002geometric}
Eleazar Eskin, Andrew Arnold, Michael Prerau, Leonid Portnoy, and Sal Stolfo.
\newblock A geometric framework for unsupervised anomaly detection.
\newblock In \emph{Applications of data mining in computer security}, pages
  77--101. Springer, 2002.

\bibitem[French et~al.(2019)French, Kokoszka, Stoev, and
  Hall]{french2019quantifying}
Joshua French, Piotr Kokoszka, Stilian Stoev, and Lauren Hall.
\newblock Quantifying the risk of heat waves using extreme value theory and
  spatio-temporal functional data.
\newblock \emph{Computational Statistics \& Data Analysis}, 131:\penalty0
  176--193, 2019.

\bibitem[Fu et~al.(2005)Fu, Hu, and Tan]{fu2005similarity}
Zhouyu Fu, Weiming Hu, and Tieniu Tan.
\newblock Similarity based vehicle trajectory clustering and anomaly detection.
\newblock In \emph{ICIP}, volume~2, pages II--602. IEEE, 2005.

\bibitem[Gama et~al.(2014)Gama, {\v{Z}}liobait{\.e}, Bifet, Pechenizkiy, and
  Bouchachia]{gama2014survey}
Jo{\~a}o Gama, Indr{\.e} {\v{Z}}liobait{\.e}, Albert Bifet, Mykola Pechenizkiy,
  and Abdelhamid Bouchachia.
\newblock A survey on concept drift adaptation.
\newblock \emph{ACM computing surveys (CSUR)}, 46\penalty0 (4):\penalty0 44,
  2014.

\bibitem[Garcia-Teodoro et~al.(2009)Garcia-Teodoro, Diaz-Verdejo,
  Maci{\'a}-Fern{\'a}ndez, and V{\'a}zquez]{garcia2009anomaly}
Pedro Garcia-Teodoro, Jesus Diaz-Verdejo, Gabriel Maci{\'a}-Fern{\'a}ndez, and
  Enrique V{\'a}zquez.
\newblock Anomaly-based network intrusion detection: Techniques, systems and
  challenges.
\newblock \emph{computers \& security}, 28\penalty0 (1-2):\penalty0 18--28,
  2009.

\bibitem[Goldstein and Uchida(2016)]{goldstein2016comparative}
Markus Goldstein and Seiichi Uchida.
\newblock A comparative evaluation of unsupervised anomaly detection algorithms
  for multivariate data.
\newblock \emph{PloS one}, 11\penalty0 (4):\penalty0 e0152173, 2016.

\bibitem[G{\"o}rnitz et~al.(2013)G{\"o}rnitz, Kloft, Rieck, and
  Brefeld]{gornitz2013toward}
Nico G{\"o}rnitz, Marius Kloft, Konrad Rieck, and Ulf Brefeld.
\newblock Toward supervised anomaly detection.
\newblock \emph{Journal of Artificial Intelligence Research}, 46:\penalty0
  235--262, 2013.

\bibitem[Guggilam et~al.(2019)Guggilam, Zaidi, Chandola, and Patra]{INCAD}
Sreelekha Guggilam, S.~M.~Arshad Zaidi, Varun Chandola, and Abani Patra.
\newblock Bayesian anomaly detection using extreme value theory.
\newblock \emph{arXiv preprint arXiv:1905.12150}, 2019.

\bibitem[He et~al.(2003)He, Xu, and Deng]{he2003discovering}
Zengyou He, Xiaofei Xu, and Shengchun Deng.
\newblock Discovering cluster-based local outliers.
\newblock \emph{Pattern Recognition Letters}, 24\penalty0 (9-10):\penalty0
  1641--1650, 2003.

\bibitem[Hjort et~al.(2010)Hjort, Holmes, Mueller, and Walker]{Hjort:2010}
N.~Hjort, C.~Holmes, P.~Mueller, and S.~Walker.
\newblock \emph{Bayesian Nonparametrics: Principles and Practice}.
\newblock Cambridge University Press, Cambridge, UK, 2010.

\bibitem[Jiang et~al.(2015)Jiang, Castner, Hewner, and Chandola]{Jiang:2015}
Jialiang Jiang, Jessica Castner, Sharon Hewner, and Varun Chandola.
\newblock Improving quality of care using data science driven methods.
\newblock In \emph{{UNYTE} Scientific Session - Hitting the Accelerator: Health
  Research Innovation through Data Science}, 2015.

\bibitem[Jiang and Gruenwald(2006)]{jiang2006research}
Nan Jiang and Le~Gruenwald.
\newblock Research issues in data stream association rule mining.
\newblock \emph{ACM Sigmod Record}, 35\penalty0 (1):\penalty0 14--19, 2006.

\bibitem[Kruegel and Vigna(2003)]{kruegel2003anomaly}
Christopher Kruegel and Giovanni Vigna.
\newblock Anomaly detection of web-based attacks.
\newblock In \emph{Proc. of the 10th ACM conference on Computer and
  communications security}, pages 251--261, 2003.

\bibitem[Lavin and Ahmad(2015)]{numentapressrelease}
Alexander Lavin and Subutai Ahmad.
\newblock Evaluating real-time anomaly detection algorithms--the numenta
  anomaly benchmark.
\newblock In \emph{2015 IEEE 14th International Conference on Machine Learning
  and Applications (ICMLA)}, pages 38--44. IEEE, 2015.

\bibitem[Neal(2000)]{Neal:2000}
Radford~M. Neal.
\newblock Markov chain sampling methods for dirichlet process mixture models.
\newblock \emph{Journal of Computational and Graphical Statistics}, 9\penalty0
  (2):\penalty0 249--265, 2000.

\bibitem[Rasmussen(2000)]{Rasmussen:2000}
Carl~Edward Rasmussen.
\newblock The infinite gaussian mixture model.
\newblock In \emph{In Advances in Neural Information Processing Systems 12},
  pages 554--560. MIT Press, 2000.

\bibitem[Shotwell and Slate(2011)]{Shotwell:2011}
Matthew~S. Shotwell and Elizabeth~H. Slate.
\newblock Bayesian outlier detection with dirichlet process mixtures.
\newblock \emph{Bayesian Anal.}, 6\penalty0 (4):\penalty0 665--690, 12 2011.

\bibitem[Siffer et~al.(2017)Siffer, Fouque, Termier, and Largouet]{Siffer:2017}
Alban Siffer, Pierre-Alain Fouque, Alexandre Termier, and Christine Largouet.
\newblock Anomaly detection in streams with extreme value theory.
\newblock In \emph{Proc. of the 23rd ACM SIGKDD International Conference on
  Knowledge Discovery and Data Mining}, pages 1067--1075, 2017.
\newblock ISBN 978-1-4503-4887-4.

\bibitem[Teh et~al.(2006)Teh, Jordan, Beal, and Blei]{Teh:2006}
Yee~Whye Teh, Michael~I. Jordan, Matthew~J. Beal, and David~M. Blei.
\newblock Hierarchical dirichlet processes.
\newblock \emph{Journal of the American Statistical Association}, 101\penalty0
  (476):\penalty0 1566--1581, 2006.

\bibitem[Varadarajan et~al.(2017)Varadarajan, Subramanian, Ahuja, Moulin, and
  Odobez]{Varadarajan:2017}
J.~Varadarajan, R.~Subramanian, N.~Ahuja, P.~Moulin, and J.~M. Odobez.
\newblock Active online anomaly detection using dirichlet process mixture model
  and gaussian process classification.
\newblock In \emph{2017 {IEEE} WACV}, pages 615--623, 2017.

\end{thebibliography}
\end{document}